\documentclass[letterpaper]{article} 
\usepackage{aaai18}
\usepackage{times}
\usepackage{helvet}
\usepackage{courier}
\usepackage{url}
\usepackage{graphicx}
\newtheorem{THEOREM}{Theorem}[section]
\newenvironment{theorem}{\begin{THEOREM} \hspace{-.85em} {\bf :} }%
                        {\end{THEOREM}}
\newtheorem{LEMMA}[THEOREM]{Lemma}
\newenvironment{lemma}{\begin{LEMMA} \hspace{-.85em} {\bf :} }%
                      {\end{LEMMA}}
\newtheorem{COROLLARY}[THEOREM]{Corollary}
\newenvironment{corollary}{\begin{COROLLARY} \hspace{-.85em} {\bf :} }%
                          {\end{COROLLARY}}
\newtheorem{PROPOSITION}[THEOREM]{Proposition}
\newenvironment{proposition}{\begin{PROPOSITION} \hspace{-.85em} {\bf :} }%
                            {\end{PROPOSITION}}
\newtheorem{DEFINITION}[THEOREM]{Definition}
\newenvironment{definition}{\begin{DEFINITION} \hspace{-.85em} {\bf :} \rm}%
                            {\end{DEFINITION}}
\newtheorem{CLAIM}[THEOREM]{Claim}
\newenvironment{claim}{\begin{CLAIM} \hspace{-.85em} {\bf :} \rm}%
                            {\end{CLAIM}}
\newtheorem{EXAMPLE}[THEOREM]{Example}
\newenvironment{example}{\begin{EXAMPLE} \hspace{-.85em} {\bf :} \rm}%
                            {\end{EXAMPLE}}
\newtheorem{REMARK}[THEOREM]{Remark}
\newenvironment{remark}{\begin{REMARK} \hspace{-.85em} {\bf :} \rm}%
                            {\end{REMARK}}

\newcommand{\thm}{\begin{theorem}}
\newcommand{\lem}{\begin{lemma}}
\newcommand{\pro}{\begin{proposition}}
\newcommand{\dfn}{\begin{definition}}
\newcommand{\rem}{\begin{remark}}
\newcommand{\xam}{\begin{example}}
\newcommand{\cor}{\begin{corollary}}

\newcommand{\ethm}{\end{theorem}}
\newcommand{\elem}{\end{lemma}}
\newcommand{\epro}{\end{proposition}}
\newcommand{\edfn}{\bbox\end{definition}}
\newcommand{\erem}{\bbox\end{remark}}
\newcommand{\exam}{\bbox\end{example}}
\newcommand{\ecor}{\end{corollary}}

\newcommand{\beqn}{\begin{equation}}
\newcommand{\eeqn}{\end{equation}}

\newcommand{\bbox}{\vrule height7pt width4pt depth1pt}

\newcommand{\clm}{\begin{claim}}
\newcommand{\eclm}{\end{claim}}

\newcommand{\sat}{\models}

\newcommand{\union}{\cup}
\newcommand{\inter}{\cap}

\renewcommand{\phi}{\varphi}

\newcommand{\E}{{\cal E}}
\newcommand{\F}{{\cal F}}

\newcommand{\K}{{\cal K}}

\newcommand{\R}{{\cal R}}

\newcommand{\U}{{\cal U}}
\newcommand{\V}{{\cal V}}

\newcommand{\ol}{\setlength{\itemsep}{0pt}\begin{enumerate}}
\newcommand{\eol}{\end{enumerate}\setlength{\itemsep}{-\parsep}}
\newcommand{\ul}{\setlength{\itemsep}{0pt}\begin{itemize}}
\newcommand{\dl}{\setlength{\itemsep}{0pt}\begin{description}}
\newcommand{\edl}{\end{description}\setlength{\itemsep}{-\parsep}}
\newcommand{\eul}{\end{itemize}\setlength{\itemsep}{-\parsep}}

\newcommand{\cF}{{\cal F}}

\newcommand{\cS}{{\cal S}}

\newcommand{\BS}{B^{\scriptscriptstyle \cS}}

\newcommand{\commentout}[1]{}

\newcommand{\bi}{\begin{itemize}}
\newcommand{\ei}{\end{itemize}}
\newcommand{\be}{\begin{enumerate}}
\newcommand{\ee}{\end{enumerate}}

 \newcommand{\intension}[2]{[\![ #2 ]\!]_{ #1}}
 
 \newcommand{\BT}{\mbox{{\it BT}}}
 \newcommand{\PL}{\mbox{{\it PL}}}
 \newcommand{\TrH}{\mbox{{\it TH}}}
 \newcommand{\TS}{\mbox{{\it TS}}}
 
 \newcommand{\CL}{\mbox{{\it CL}}}
 
  \newcommand{\LT}{\mbox{{\it LT}}}
  \newcommand{\SD}{\mbox{{\it SD}}}
 \newcommand{\ST}{\mbox{{\it ST}}}
 \renewcommand{\BS}{\mbox{{\it BS}}}
 \newcommand{\mr}{\mbox{{\it db}}}
 \newcommand{\cost}{\mbox{{\it c}}}
  \newcommand{\pw}{\mbox{{\it pw}}}
 \newcommand{\util}{\mathbf{u}}
\newcommand{\JS}{\mbox{{\it JS}}}
\newcommand{\ACT}{A}
\newcommand{\act}{a}
 \newcommand{\JP}{\mbox{{\it JP}}}
     \newcommand{\BP}{\mbox{{\it BP}}}
          \newcommand{\ag}{\mathbf{ag}}
          \newcommand{\INTEND}{\mathit{INTEND}}
\renewcommand{\citeyear}{\shortcite}
\renewcommand{\S}{{\cal S}}
\newcommand{\REF}{{\mathit{REF}}}
\newcommand{\shortv}{\commentout}
\newcommand{\fullv}[1]{#1}
\newcommand{\MX}{N}
\usepackage{todonotes} 
\newcommand{\mtodo}[2][]
{}{}
\newcommand{\jtodo}[2][]
{}{}

\title{Towards Formal Definitions of Blameworthiness, Intention, and
Moral Responsibility}

\author{Joseph Y. Halpern\\
Dept. of Computer Science\\
Cornell University\\
Ithaca, NY 14853\\
halpern@cs.cornell.edu
\And
Max Kleiman-Weiner\\
    Brain and Cognitive Sciences\\
    Massachusetts Institute of Technology\\
Cambridge, MA 02139\\
maxkw@mit.edu}

\vspace{-10em} 

\begin{document}

\maketitle

 \begin{abstract}
     We provide formal definitions of \emph{degree of blameworthiness} and
     \emph{intention} relative to an \emph{epistemic state} (a probability
          over causal models and a utility function on outcomes).  
     These, together with a definition of actual causality, 
     provide the key ingredients for moral responsibility judgments.
          We show that these definitions give insight into commonsense
     intuitions in a variety of puzzling cases from the literature.
 \end{abstract}

\section{Introduction}
The need for judging \emph{moral responsibility}
arises both in ethics and in law.  In an era of autonomous vehicles
and, more generally, autonomous AI agents that interact with or
on behalf of people, the issue has now become 
relevant to AI as well.
\commentout{
To repeat the standard trolley problem
\cite{Foot67}, suppose that a runaway trolley is heading down the
tracks.  There are 5 people tied up on the track, who cannot move.  If
the trolley continues, it will kill all 5 of them.  While you cannot
stop the trolley, you can pull a lever, which will divert it to a
side track.  Unfortunately, there is a man on the side track who will
get killed if you pull the lever.  What is appropriate thing to do here?
What is your degree of moral responsibility for the outcome if you
do/do not pull the lever.}
We will clearly need to imbue AI agents with some means for
evaluating moral responsibility.
There is general agreement that a definition of moral responsibility
will require integrating causality, some notion of \emph{blameworthiness}, and
\emph{intention} \cite{Cushman15,malle2014theory,Weiner95}.
Previous work has provided formal accounts of causality
\cite{Hal48}; in this paper, we provide formal
definitions of blameworthiness and intention in the same vein.

These notions are notoriously difficult to define carefully. The
well-known \emph{trolley problem} \cite{Thomson85} illustrates some of
them: Suppose that a runaway trolley is headed towards five people who
will not be able to get out of the train's path in time. If the
trolley continues, it will kill all five of them. An agent $\ag$ is near a
switchboard, and while $\ag$ cannot stop the trolley, he can pull a lever
which will divert the trolley to a side track. Unfortunately, there is
a single man on the side track who will be killed if $\ag$ pulls the
lever.

Most people agree that it is reasonable for $\ag$ to pull the
lever. But now consider a variant of the trolley problem known as
\emph{loop} \cite{Thomson85}, where instead of the side track going
off in a different direction altogether, it rejoins the main track
before where the five people are tied up. Again, there is someone on
the side track, but this time $\ag$ knows that hitting the man on
the loop will stop the train before it hits the five people on the
main track. How morally responsible is $\ag$ for the death of the
man on the side track if he pulls the lever? Should the answer be
different in the loop version of the problem? Pulling the lever in the
loop condition is typically judged as less morally permissible than
in the condition without a loop \cite{Mikhail07}.
The definitions given here take as their starting point the
\emph{structural-equations} 
 framework used by Halpern and Pearl
\citeyear{HP01b}
(HP from now on) in defining 
causality.  This framework allows us to model counterfactual
statements like
``outcome $\phi$ would have occurred if agent $\ag$ had performed
$\act'$ rather than $\act$''.  Evaluating such counterfactual statements
is the key to defining \emph{intention} and \emph{blameworthiness},
which are significant components of moral responsibility,
just as it is for defining actual causation. 
Since we do not assume that actions lead deterministically to
outcomes, we need to have a probability on the effects of actions.
Once we consider causal models augmented with probability, we can 
to define $\ag$'s \emph{degree
   of blameworthiness}; rather than $\ag$ either being
blameworthy or not for an outcome, he is only blameworthy to 
some degree (a number in [0,1]).
If we further assume that the agent is an expected-utility maximizer, 
and augment the framework with a utility function, we can also define
\emph{intention}.  Roughly 
speaking, an agent who performs action $\act$ intends outcome $\phi$ if he
would not have done $\act$ if $\act$ had no impact on whether
$\phi$ occurred.
(We use the assumption that the agent is an expected-utility maximizer
to determine what the agent would have done if $\act$ had no impact on
$\phi$.)


\commentout{
Moral responsibility has been studied in philosophy since the 
time of the Greeks (see \cite{Eshleman14} for an
overview). \mtodo{possible PC issue here... what about the
  Indians/Chinese, etc.}
Issues
related to moral responsibility and intention have also been of
interest in AI for a while.  In particular, since in many
accounts of moral responsibility an agent's beliefs, desires, and
intentions play a key role, work in AI on beliefs, desires, and
intentions (BDI) (see, e.g., \cite{RG95}) becomes relevant.  The fact
that there are many implementations of BDI systems (see
en.wikipedia.org/wiki/Belief\%E2
for a recent list) suggests that the computational concerns might be
addressable.  However, the work on BDI does not consider formal
definitions of moral responsibility.  
Cohen and Levesque \citeyear{CL90} provide a formal definition of
intention in a modal logic setting.  
More recently, Kleiman-Weiner et al.~\shortcite{KWG15} also gave a
definition of intention, and applied it to moral decision making 
Work on causality is also relevant to moral responsibility; indeed, the
Halpern-Pearl (HP) framework using structural equations (see
\cite{Hal48} for recent relevant work and additional references) will
form the basis for the definitions here.  
  Chockler and Halpern \citeyear{ChocklerH03} have defined
a notion of responsibility based on causality,
but their definitions, while relevant to the work discussed here,
explicitly do not attempt to get at moral issues.}
\fullv{
The rest of this paper is organized as follows.
In Section~\ref{sec:causality}, we review the structural-equations
framework and the HP definition of causality.  In
Section~\ref{sec:blame}, we define degree of blameworthiness.  In
Section~\ref{sec:intention}, we define intention.
We discuss computational issues
in Section~\ref{sec:complexity}.
There is a huge literature on moral responsibility and intention; we
discuss the most relevant related work in Section~\ref{sec:related},
and conclude in Section~\ref{sec:conclusion}.
}

\commentout{
\section{Moral responsibility issues}\label{sec:issues}
Suppose that an agent $\ag$ performs an action $\act$ that leads to an outcome
$\phi$.  In this section, I consider what
should go into determining the degree to which $\ag$ is morally responsible for
$\phi$.

\paragraph{Causality:}
A relatively uncontroversial necessary condition for moral
responsibility is causality:
For $\ag$ to be morally responsible for $\phi$, 
action $\act$ has to be (part of) a cause for $\phi$.  There
are many definitions of causality in the literature.  For 
definiteness, in this paper, I use the Halpern-Pearl (HP) definition
\cite{HP01b}, as most recently modified by Halpern \citeyear{Hal47}.
(See \cite{Hal48} for more discussion.)  \mtodo{switches to intention, but I thought this paragraph was about moral responsibility?}
\jtodo{I don't understand this comment; the paragraph above and the one below are about causality, not intention.}
\mtodo{the paragraph below mentions intention out of the blue: ``needed to define intention''}
I do this for several reasons.  First, the HP definition arguably
handles most examples better than other definitions.  Second, it can
be extended easily to handle defaults and normality \cite{HH11} (see also
Section~\ref{??}).  Finally, the structural-equations framework used
in the HP definition plays an important role in defining other notions
(in particular, the counterfactuals needed to define intention).
That said, 
the HP definition could be replaced by another
definition of causality (e.g.,
\cite{GW07,Hall07,HP01b,hitchcock:99,Hitchcock07,Woodward03,wright:88}). 

\paragraph{Beliefs and Desires:}  
For $\ag$ to be morally responsible for an outcome $\phi$, he must be
viewed as deserving of blame (or praise) for $\phi$.  In determining
blameworthiness,  
it does not suffice that $\ag$'s action causes $\phi$; among other
things, $\ag$ has to believe that his action 
will cause $\phi$, or at least has some likelihood of causing $\phi$.  If
$\ag$ had no idea that his action would cause $\phi$, then we do not in
general want to blame $\ag$ for $\phi$ (but see
below). \mtodo{might want to make it stronger by saying if $\ag$ had
  no way to know -- that way it won't be confused with negligence.}
\jtodo{There's no planning here, so no easy way to say "had no way to
  know"; I'm identifying "had no idea" with "puts probability 0 on"}
But 
what if $\ag$ only 
believes that his action would cause the outcome with probability
$1/5$?  Is he morally responsible in that case?  It does not seem
reasonable to set a probability threshold above which we say that the
agent is morally responsible and below which he is not.  The solution
adopted here is to talk about the \emph{degree} of blameworthiness.

But we do not always want to consider just the agent's subjective
probability here.  For example, suppose that the agent had several
bottles of beer, goes for a drive, and runs over a pedestrian.  
The agent may well have believed that the probability 
that his driving would cause an accident was
low.  In cases like this, the courts typically consider the
probability that a ``reasonable person'' would have regarding the
probability of an accident, not the agent's probability.   In the
sequel, I call this \emph{society's probability}.

One situation where we do not want to use society's probability is in
the case of insanity.  The so-called \emph{M'Naghten test} \cite{Elliott96}
essentially says that an agent should not be held
morally responsible in the eyes of the law if he did not know ``the
nature and quality of the act he was doing''.  Certainly one part of
knowing the nature and quality of an act involves having a reasonably accurate
estimate of the probability that it will cause a particular outcome.
Thus, if there is independent reason to believe that an agent $\ag$ is
sufficiently impaired so that the M'Naghten test applies, then $\ag$
should not be held morally responsible; specifically, we should not
apply society's probability. \mtodo{I think this is only partially
  true... do we think insane people just don't know the estimates of
  different probabilities or rather do we think that insane people
  fail in a computational sense rather than an epistemic sense of
  failing to act as utility maximizing rational agents?}\jtodo{I
  suspect it's all of the above.  Typically I don't think of the
  problem with insane people as being computational.  Rather, they
  simply have different probabilities and/or utilities than the rest
  of society, but we don't blame them for it.} 

Probability alone is not enough to determine how good an action is; we
need to know the agent's desires.  We describe the agent's desires here
in terms of a utility function, and assume that the agent is trying to
maximize expected utility.  This is a significance assumption, although
it is relatively standard in the literature.  We discuss alternative
approaches in Section~\ref{sec:conclusion}. \mtodo{This would be a
  good place to cite my previous paper too -- in particular we should
  clarify that in this work we will assume that an agents desires are
  totally known and thus do not need to be inferred from
  action.}\jtodo{Good point.}

\paragraph{Intention:}
Two types of intention have been considered (see, e.g., \cite{CL90}):
(1) whether agent $\ag$ intended to perform action $\act$ (perhaps it was
an accident) and (2) did $\ag$ (when performing $\act$) intend outcome
$\phi$ (perhaps $\phi$ was an unintended side-effect of $\act$, which was
actually performed to bring about outcome $o'$). 
\mtodo{in my work I call these ``intend-to'' and ``intend-that''} 
There are formal 
definitions of moral responsibility (e.g., \cite{BH12}) that take 
intending to perform the action to be a necessary condition for moral
responsibility.  With regard to (2), many papers have argued that intending the
outcome is an important aspect of moral responsibility (see, e.g.,
Waldmann, Nagel, and Weigmann's \citeyear{WNW12} review); there is
also evidence showing that people's judgments of moral responsibility
is influenced by intentionality.  The issue is perhaps best understood
in the context of the well-studied \emph{trolley problem} \cite{Foot67}.

\xam\label{xam:trolley}
In the standard trolley problem, there are five people tied up on the
track, who cannot move.  If the trolley continues, it will kill all five
of them.  While $\ag$ cannot stop the trolley, $\ag$ can pull a lever,
which will divert the trolley to a 
side track.  Unfortunately, there is a man on the side track who will
get killed if $\ag$ pulls the lever.  In this case, most people agree
that it is reasonable for $\ag$ to pull the lever.
But now consider a variant of
the trolley problem known as \emph{loop} \cite{Thomson85}, where
  instead of the side track going off in a different direction
  altogether, it rejoins the main track before where the five people
  are tied up.  Again, there is someone on the side track, but
  this time the agent knows that hitting the agent will stop the
  train before it hits the five people on the main track.  Pulling the
  lever in the loop   condition is viewed as less 
  morally permissible for killing the man than in the condition without a loop
  \cite{Mikhail07}.  One explanation for this is that an agent who pulls
  the lever in the loop condition intends to kill the man on the
  side track; he would not have pulled the lever had the man not been
  there.  On the other hand, in the more standard side track condition, the
  agent who pulled the lever did not intend to kill the man; he would
  have pulled the lever even if the man had not been there.
  A related story considers a hospital setting where two people are in
    need of kidneys, a third is in need of a heart, a fourth is in
    need of lungs, and a fifth is in
  need of a liver.  All will die unless they get the appropriate transplant.
  A perfectly healthy man comes in.  Most people
  find it completely inappropriate to kill him in order to save the
  five people, although, just as in the trolley and the loop problem,
  we are killing one person to save five.
\exam


While intention seems to affect people's judgments of
moral responsibility, neither form of intention is \emph{necessary}
for moral responsibility.   Recall the drunk driver who
unintentionally runs over a pedestrian.  Most people would 
consider the driver morally responsible although he certainly did not intend
to perform the action of running over the pedestrian (nor did he
intend the outcome). 
\commentout{
The following example shows that even
if you do not intend an outcome, you can be morally responsible for it 
\mtodo{I think the language preceding this needs to clarify moral
  responsibility for an outcome and moral responsibility for an action
  (with all the outcomes).  }.


\xam\label{xam:Sibella} Suppose that Louis wants to kill his cousin
Rufus, who is 
  standing in the way of him getting an inheritance.  He knows that
  Rufus will be having lunch with Sibella, and will be sitting at a
  particular table.   He plants a bomb at that table, which
  unfortunately (from Louis' point of view) kills Sibella, but not
  Rufus.  Louis can (truthfully) claim that he did not intend to kill
  Sibella; he only wanted to kill Rufus.  Had Rufus come to lunch on his
 own, Louis would still have detonated the bomb.  Yet, we suspect that
  most people would still want to call Rufus morally responsible for
  Sibella's death, despite the fact that he didn't intend it.
\exam
}

\commentout{
With regard to the first requirement, even if an agent performs an
  action by accident, we still might want to hold him morally responsible
  if the accident was foreseeable (with some probability) and
  reasonable steps could have been taken to prevent it.  If someone
  who is drunk inadvertently hits the accelerator and runs over a
  pedestrian, we would certainly want to hold them morally
  responsible, even if they didn't not intend to hit the accelerator.
  Note that here too we will need to address the issue of whose
  beliefs should be considered.  The driver might have believed that
  his alcohol intake was so low that an accident was extremely
  unlikely; society might believe otherwise.
}

In general, we do not have enough information about an agent to
determine the agent's intentions.  Specifically, we do not know how
an agent would have acted in a setting where the outcome of his
actions would have been different.
Kleiman-Weiner et
al.~\citeyear{KWG15} provide techniques for inferring intention.  While
such techniques will no doubt be helpful when it comes to determining
intentionality for people, the problem is less difficult when it comes
to autonomous AI agents: we can look at the code and
determine what the system would have done in a different setting.
\mtodo{need to be clear here, AI will still need to infer the intentions/desires of other agents. Do you mean for humans evaluating the intentions of an AI? Do the intentions of an AI matter? or just the actions? (unless you put their intentions into their utility function like a suggest above)}
\jtodo{I'm not sure if the intentions of an AI should matter.  That's on my "todo list" of things to think about.  I was deliberately trying avoiding dealing with multi-agent systems and the problem of inferring the  intentions of others.  I think i'ts reasonable to start with single-agent systems for simplicity.}
Indeed, there may even be legal requirements that make this easy to
determine.  
  
\paragraph{Considering the alternatives:} The heart of the definition of moral
responsibility given here is a consideration of what the agent's alternatives
are.  The idea is that agent $\ag$ is not morally responsible for
outcome $\phi$ unless $\ag$ had a reasonable alternative that would not
have resulted in $\phi$.  In the basic trolley example, most people would not
hold someone who pulled the lever morally responsible for the 
death that occurred, given that the alternative would have resulted in
five deaths.  The need to consider alternatives is not new, as can
be seen from the essays in \cite{WM03}.%
\footnote{In the philosophy literature,
part of the discussion of alternatives is bound 
up with issues of determinism and free will (if the world is deterministic and
agents do not have free will, then they never could have done
otherwise).  For most of this paper, we ignore this issue, and assume
that agents always have choices (although see
Section~\ref{sec:conclusion} for some discussion).}
But  the problem remains that of providing a reasonable model of
alternative choices, and how to evaluate them, even assuming that we
have probability and utility.  
%

The basic idea is straightforward.  Agent  $\ag$ who performs $\act$ is
morally responsible for outcome $\phi$ only if there is something that a
could have done that both (a) might have resulted in $\phi$ not
happening and (b) would have given
$\ag$  a higher 
expected utility. In the trolley dilemma we don't hold $\ag$ responsible for the death of the man if he
flips the lever because the alternative would have been worse.
%
But there are some subtleties.
In particular, it's not enough that there's an action that makes
things better; it must do so in a way connected with the outcome $\phi$.
\mtodo{the example below is not the basic trolley problem and is a bit confusing since no one is on the side-track. }
\xam\label{xam:trolley1}  Consider a variant of the trolley problem, but now 
suppose that, in addition to five people dying, there may be 
significant damage to property
and there are five people on both tracks.   If $\ag$ does not pull the
lever, then five people die and there is definitely significant
property damage.  If 
he does pulls the lever, then with
probability .8, there will be no property damage, but the five people
still die.  In this case,
if $\ag$ does not pull the lever,
$\ag$ is not morally responsible for the five
deaths (no matter what he did, the five people would have died), 
but is morally responsible for
the property damage. 
The point here is that just the existence
of a better action $\act'$ is not enough.  If we are considering moral
responsibility for outcome $\phi$, The action $\act'$ must be better in
affects $\phi$. \exam


There are yet other subtleties.  For one
thing, what set of alternative actions do we consider?   For example,
there may be a wonderful alternative that the agent could have come up
with had he thought about the problem harder.  We typically do not
want to fault the agent for not coming up with a difficult-to-see
alternative, but sometimes do hold the agent responsible if he did not
consider what seems like a more obvious alternative.  Braham
and van Hees \citeyear{BH12} consider as alternatives only what they call
the \emph{eligible actions}.  What these actions are is determined
outside the model.  The eligible actions might consist of all actions
that the agent can perform, but it might also consist of that subset 
that are acceptable by society (e.g., lying might not be an
eligible action if society frowns on it). 
\mtodo{this is a complicated aside... lying is wrong because it is
  intentionally bringing about something undesirable... false belief
  see Hurka 2000. Also lying is usually too abstract to be an action
  and is used instrumentally. }
I just take the set of
eligible actions to be determined by the model, just as the
probability and utility are.  The modeler can thus choose the
appropriate set of actions, the appropriate probability (the agent's
or society's), and the appropriate utility (the agent's or
society's).  In general, it may be hard to determine what the
probability and utility should be.  \mtodo{awkward sentence around sensors}
When it comes to applying
these ideas to autonomous AI agents, we will surely build into agents
ways of determining the relevant probability (e.g., sensors and models
of other agents).
We may well require that the agent be built in such
a way that an outside observer can determine the probability that the
agent was using.  When it comes to the utility function, the system
designer will presumably build a utility function into the agent, so
we will know what it is.  Exactly how the utility function should be
defined is beyond the scope of this paper.  There is unfortunately
no agreed-upon moral code that we can use.  Indeed,
cross-cultural studies have made it clear that there are essentially
no universally agreed-upon moral rules (see \cite{WNW12} for further
discussion of this point and references).  \mtodo{see HC Barrett for
  what looks like a cultural universal in using intentions to
  distinguish between side-effects and accidents. This is highly
  relevant for our paper(s) since we've built a formal model of
  intention which might just be the *only* cultural universal.}

\commentout{
\paragraph{Retrospective judgment vs.~prospective judgments:}
Judgments of actual causality are retrospective.  We would not say
that $A$ is a cause of $B$ unless $A$ and $B$ actually happened.
Since, as we said above, we do not want to say that $\ag$, who performed
action $\act$, is morally responsible responsible for $o$ unless $\act$
was a cause of $o$, the judgment of moral responsibility is also
retrospective.   However, in the main application, we are considering
prospective judgments.  That is, we might consider whether the
autonomous vehicle should swerve right to avoid an accident when doing
so might kill the occupant.  To make this decision, we want to
consider the question of moral responsibility.  Fortunately, this turns
out to be a relatively minor problem.  Roughly speaking, we will to consider
the moral responsibility that the action would have if the action were
performed.  (Of course, there is uncertainty about whether performing the action
will end up causing the outcome; for example, swerving may not cause
the car's occupant to die. But since the model that we use for moral
responsibility involves uncertainty in any case, the adaptation is
straightforward.  Essentially, to evaluate
moral responsibility prospectively, we simply remove the requirement
that the action was a cause of the outcome (causality issues come up in
other parts of the definition in any case; see  Section~\ref{??}.
}
}

\section{Structural equations and HP causality}\label{sec:causality}


The HP approach assumes that the world is described in terms of 
variables and their values.  
Some variables have a causal influence on others. This
influence is modeled by a set of {\em structural equations}.
It is conceptually useful to split the random variables into two
sets: the {\em exogenous\/} variables, whose values are
determined by 
factors outside the model, and the
{\em endogenous\/} variables, whose values are ultimately determined by
the exogenous variables.  
We assume that there is a special endogenous variable $\ACT$ called the
\emph{action variable}; the possible values of $\ACT$ are the actions
that the agent can choose among.%
\footnote{In a more general setting
with multiple agents, each performing actions, we might have a
variable $\ACT_{\ag}$ for each agent $\ag$.  We might also consider
situations over time, where agents perform sequences of actions,
determined by a strategy, rather than just a single
action.  Allowing this extra level of generality has no impact on the
framework presented here.}

For example, in the trolley problem,
we can assume that $\ACT$ has two possible values: $A=0$ if the lever
was not pulled and $A=1$ if it was.
Which action is taken is determined by an exogenous
variable.
%
The two possible outcomes in the trolley problem are 
described by two other endogenous variables: $O_1$, which is 1 if the
five people on 
the main track die, and 0 if they don't, and $O_2$, which is 1 if the
person on the sidetrack dies, and 0 otherwise.
\fullv{
Having described how actions and outcomes can be represented as
variables we can now define causal models formally.} 
A \emph{causal model} $M$
is a pair $(\S,\F)$, where $\S$ is a \emph{signature}, 
that is, a tuple $(\U,\V,\R)$, where $\U$ is a set of
exogenous variables, $\V$ is a set 
of endogenous variables, and $\R$ associates with every variable $Y \in 
\U \union \V$ a nonempty set $\R(Y)$ of possible values for 
$Y$ (i.e., the set of values over which $Y$ {\em ranges}),
and $\F$ is a set of \emph{modifiable
structural equations}, relating the values of the variables.  Formally,
$\F$ associates with each endogenous variable $X \in \V$ a
function denoted $F_X$ such that $F_X: (\times_{U \in \U} \R(U))
\times (\times_{Y \in \V - \{X\}} \R(Y)) \rightarrow \R(X)$.
\mtodo{does this mean the formalism is limited in only considering
  actions that are caused by an exogenous variable? how would this
  work in the case of manipulation or in the rainy day example I gave
  above}
In the trolley problem as modeled above,
there are two equations: $O_1 = 1- \ACT$ (the five people die
if the agent does nothing) and $O_2 = \ACT$ (the one person on the
side track dies if the agent pulls the lever).


\fullv{
  Following Halpern and Pearl \citeyear{HP01b}, we restrict attention
  here to what are called {\em 
  recursive\/} (or {\em acyclic\/}) models.  This is the special case
where there is some total ordering $\prec$ of the endogenous variables
(the ones in $\V$) 
such that if $X \prec Y$, then $X$ is independent of $Y$, 
that is, $F_X(\ldots, y, \ldots) = F_X(\ldots, y', \ldots)$ for all $y, y' \in
\R(Y)$.
If $X \prec Y$, then the value of $X$ may affect the value of
$Y$, but the value of $Y$ cannot affect the value of $X$.
It should be clear that if $M$ is an acyclic  causal model,
then given a \emph{context}, that is, a setting $\vec{u}$ for the
exogenous variables in $\U$, there is a unique solution for all the
equations.
\fullv{
We simply solve for the variables in the order given by
$\prec$. The value of the variable that comes first in the order, that
is, the variable $X$ such that there is no variable $Y$ such that $
Y\prec X$, depends only on the exogenous variables, so $X$'s value is
immediately determined by the values of the exogenous variables.  
The values of variables later in the order can be determined once we have
determined the values of all the variables earlier in the order.}
}

\shortv{
Just as HP do, we restrict attention to 
{\em acyclic\/} causal models, where 
there is a total ordering $\prec$ of the endogenous variables
(the ones in $\V$) 
such that if $X \prec Y$, then $X$ is independent of $Y$, 
that is, $F_X(\ldots, y, \ldots) = F_X(\ldots, y', \ldots)$ for all $y, y' \in
\R(Y)$.
If $X \prec Y$, then the value of $X$ may affect the value of
$Y$, but the value of $Y$ cannot affect the value of $X$.
It should be clear that if $M$ is an acyclic  causal model,
then given a \emph{context}, that is, a setting $\vec{u}$ for the
exogenous variables in $\U$, there is a unique solution for all the
equations.}

Given a causal model $M = (\cS,\cF)$, a
vector $\vec{X}$ of distinct variables in $\V$, and a vector $\vec{x}$ 
of values for the variables in
$\vec{X}$, the causal model $M_{\vec{X} \gets \vec{x}}$
is identical to $M$, except that the
equation for the variables $\vec{X}$ in $\cF$ is replaced by $\vec{X} =
\vec{x}$. 
Intuitively, this is the causal model that results when the variables in
$\vec{X}$ are set to $\vec{x}$ by some external action
that affects only the variables in $\vec{X}$
(and overrides the effects of the causal equations).

To define causality carefully, it is useful to have a language to reason
about causality.
Given a signature $\S = (\U,\V,\R)$, a \emph{primitive event} is a
formula of the form $X = x$, for  $X \in \V$ and $x \in \R(X)$.  
A {\em causal formula (over $\S$)\/} is one of the form
$[Y_1 \gets y_1, \ldots, Y_k \gets y_k] \phi$,
where
\fullv{\begin{itemize}\item}
$\phi$ is a Boolean
combination of primitive events,
\fullv{\item} $Y_1, \ldots, Y_k$ are distinct variables in $\V$, and
\fullv{\item} $y_i \in \R(Y_i)$.
\fullv{\end{itemize}}
Such a formula is
abbreviated
as $[\vec{Y} \gets \vec{y}]\phi$.
The special
case where $k=0$
is abbreviated as
$\phi$.
Intuitively,
$[Y_1 \gets y_1, \ldots, Y_k \gets y_k] \phi$ says that
$\phi$ would hold if
$Y_i$ were set to $y_i$, for $i = 1,\ldots,k$.

A pair $(M,\vec{u})$ consisting of a causal model and a
context is called a \emph{causal setting}.
A causal formula $\psi$ is true or false in a causal setting. 
As in HP, $(M,\vec{u}) \sat \psi$  if
the causal formula $\psi$ is true in the 
causal setting  $(M,\vec{u})$.
The $\sat$ relation is defined inductively.
$(M,\vec{u}) \sat X = x$ if
the variable $X$ has value $x$
in the
unique (since we are dealing with acyclic models) solution
to
the equations in
$M$ in context $\vec{u}$
(i.e., the
unique vector
of values for the exogenous variables that simultaneously satisfies all
equations 
in $M$ 
with the variables in $\U$ set to $\vec{u}$).
The truth of conjunctions and negations is defined in the standard way.
Finally, 
$(M,\vec{u}) \sat [\vec{Y} \gets \vec{y}]\phi$ if 
$(M_{\vec{Y} \gets \vec{y}},\vec{u}) \sat \phi$.

\shortv{
  In the full paper,%
    \footnote{Available at www.cs.cornell.edu/home/halpern/moralresp.pdf.}
  the HP definition of causality is given.
   The details are not necessary for understanding the remaining
   definitions. Indeed, the HP definition could be replaced by another
   definition of causality (e.g.,
   \cite{GW07,Hall07,HP01b,hitchcock:99,Hitchcock07,Woodward03,wright:88}).}

\fullv{
The HP definition of causality, like many others, is based on
counterfactuals.  The idea is that $A$ is a cause of $B$ if,
had $A$ not occurred (although it did), then $B$ would not have occurred.
But there are many examples showing that this naive definition will not
quite work.  For example, suppose that Suzy throws a rock at a bottle,
and shatters it.  Billy is waiting in the wings with his rock; if Suzy
hadn't thrown her rock, then Billy would have thrown his, and shattered
the bottle.  We would like to say that Suzy is a cause of the bottle
shattering, but if Suzy hadn't thrown, then the bottle would have
shattered anyway (since Billy would have thrown his rock).  The
definition is intended to deal with this example (and many others).
While the HP definition has been shown to work well, it could be
replaced by another definition of causality
based on counterfactuals
(e.g., \cite{GW07,Hall07,HP01b,hitchcock:99,Hitchcock07,Woodward03,wright:88})
without affecting the remaining definitions in the paper.

In the definition, what can be a cause is a conjunction $X_1 = x_1
\land \ldots \land X_k=x_k$ of primitive events
(where $X_1,\ldots,X_k$ are distinct variables),
typically abbreviated
as $\vec{X} = \vec{x}$; what can be caused is an arbitrary Boolean
combination $\phi$ of primitive events.  
\dfn\label{actcaus}
$\vec{X} = \vec{x}$ is an {\em actual cause of $\phi$ in
$(M, \vec{u})$ \/} if the following
three conditions hold:
\begin{description}
\item[{\rm AC1.}]\label{ac1} $(M,\vec{u}) \sat (\vec{X} = \vec{x})$ and 
$(M,\vec{u}) \sat \phi$.
\item[{\rm AC2.}]\label{ac2}
There is a set $\vec{W}$ of variables in $\V$
and a setting $\vec{x}'$ of the variables in $\vec{X}$ such that 
if  $(M,\vec{u}) \sat \vec{W} = \vec{w}$, then
$$(M,\vec{u}) \sat [\vec{X} \gets \vec{x}',
\vec{W} \gets \vec{w}]\neg \phi.$$
\item[{\rm AC3.}] \label{ac3}
$\vec{X}$ is minimal; no subset of $\vec{X}$ satisfies
conditions AC1 and AC2.
\label{def3.1}  
\end{description}
If $\vec{X} = \vec{x}$
is a cause
of $\phi$ in $(M,\vec{u})$ and $X=x$ is a
conjunct of $\vec{X}= \vec{x}$, then $X=x$ is \emph{part of a cause of
  $\phi$ in $(M,u)$}.
\end{definition}

AC1 just says that $\vec{X}=\vec{x}$ cannot
be considered a cause of $\phi$ unless both $\vec{X} = \vec{x}$ and
$\phi$ actually happen.  AC3 is a minimality condition that ensures
that only those elements of 
the conjunction $\vec{X}=\vec{x}$ that are essential are
considered part of a cause; inessential elements are pruned.
Without AC3, if dropping a lit cigarette is a cause of a fire
then so is dropping the cigarette and sneezing.
AC2 is the core of the definition.
If we ignore $\vec{W}$, it is essentially the standard
counterfactual definition: if $\vec{X}$ is set to some value
$\vec{x}'$ other than its actual value $\vec{x}$, then $\phi$ would
not have happened.  As we observed, this is not enough to deal with
the case of Billy waiting in the wings.  The actual definition allows
us to consider what happens if Suzy doesn't throw, while keeping fixed
the fact that Billy didn't throw (which is what happened in the actual
world); that is, if the causal model includes binary variables%
\footnote{A variable is \emph{binary} if it has two possible values.}
$\ST$ (for
Suzy throws), $\BT$ (for Billy throws) and $\BS$ (for bottle
shatters), with the equation $\BT = 1-\ST$ (Billy throws exactly if
Suzy doesn't) and $\BS = \ST \lor \BT$ (the bottle shatters if either
Billy or Suzy throws), and $\vec{u}$ is the context where Suzy throws, 
then we have $(M,u) \sat [ST \gets 0, \BT \gets 0](\BS = 0)$, so AC2
holds.  
}

\section{Degree of blameworthiness}\label{sec:blame}

We now apply this formal language to study blameworthiness. 
For agent $\ag$ to be morally responsible for an outcome $\phi$, he
must be viewed as deserving of blame for $\phi$.
Among other things, for $\ag$ to be deserving of blame, he
must have placed some likelihood (before acting) on the possibility
that performing $\act$ would affect $\phi$. If $\ag$ did not believe it was
possible for $\act$ to affect $\phi$, then in general we do not want to
blame $\ag$ for $\phi$ (assuming that 
$\ag$'s beliefs are reasonable; see below).

In general, an agent has uncertainty regarding the structural
equations that characterize a causal model and about the context.
This uncertainty is characterized
by a probability distribution $\Pr$ on a
set $\K$ of causal settings.%
\footnote{Chockler and Halpern \citeyear{ChocklerH03} also used such a
  probability to define a notion of \emph{degree of blame}.}
\commentout{
The ``roughly speaking'' here is due to the fact that the causal
setting also determines the value of the action variable $\ACT$.  But when
evaluating blameworthiness, we want to treat the agent as being
able to freely choose his action $\act$ rather than
the choice
being determined by 
the values of  exogenous variables.
To make this precise, assume that there is a special exogenous variable 
$U_{\ACT}$ that affects only $\ACT$ 
and is the only exogenous variable that directly affects $\ACT$
(i.e., $U_{\ACT}$ appears only in the equation for $\ACT$ and is the
only exogenous variable that appears in the equation for $\ACT$).  Intuitively,
$U_{\ACT}$ captures the agent's state of mind insofar as performing
$\ACT$.  $\ACT$ may be affected by other endogenous variables.
For example, the agent's decision may be determined by (among other things)
genetic factors and environmental factors.
This lets us model the extent to which $\ag$'s action was
beyond $\ag$'s control.

Let a \emph{reduced 
  context} consist of the settings of all exogenous variables other
than $U_{\ACT}$.  Note that the truth of formulas of the form $[\ACT \gets
  \act]\phi$ (and, more generally, of formulas of the form $[\ACT \gets \act,
  \vec{Y} \gets \vec{y}]\phi$) depends only on the reduced context.
  For ease of exposition, we write $(M,\vec{u}) \sat \psi$ even if
 $\vec{u}$ is a reduced context, as long as it is well defined.
}
Let $\K$ consist of causal settings $(M,\vec{u})$, and let $\Pr$ be a
probability measure on $\K$.
$\Pr$ should be
thought of as describing the probability \emph{before} the action is
performed.  For ease of exposition, we assume that all the models in
$\K$ have the same signature (set of endogenous and exogenous variables).
We assume that an agent's preferences are characterized by a utility
function $\util$ 
on \emph{worlds}, where a \emph{world} is a complete assignment to
the endogenous 
variables.
%
Thus, an \emph{epistemic state} for an agent $\ag$ consists of a tuple
$\E = (\Pr,\K,\util)$.
\commentout{
We normalize $\util$ so that the utility of the best world
is 1 and the utility of the worst world is 0
(for reasons explained below).%
\footnote{
  As is well known, for an agent whose preference order on
  actions is determined by expected utility, so that he prefers $\act$
  to $\act'$ iff the expected utility of $\act$ is higher than that of
    $\act'$, replacing a utility function $\util$ by a utility function $\util'$
  that is an affine transformation of $\util$ (i.e., $\util' =
  r_1\util + r_2$ for 
  some real-valued constants $r_1$ and $r_2$) has no impact on the
  agent's preferences.  Thus, we can assume without loss of generality
  that $\util$ is normalized in this way.
}
}

Given an epistemic state for an agent $\ag$, we can determine the extent to
which $\ag$ performing action $\act$
affected, or 
made a difference, to an outcome $\phi$ (where $\phi$ can be an arbitrary
Boolean combination of primitive events).  Formally, we compare $\act$ to all
other actions $\act'$ that $\ag$ could have performed.
Let $\intension{\K}{\phi} = \{(M,\vec{u}) \in \K: (M,\vec{u}) \sat
\phi\}$; that is, $\intension{\K}{\phi}$ consists of all causal
settings in $\K$ where $\phi$ is true. 
Thus, $\Pr(\intension{\K}{[\ACT = \act]\phi})$ is the probability that
performing action $\act$ results in $\phi$. 
Let $$\delta_{\act,\act',\phi} = \max(0, \Pr(\intension{\K}{[\ACT = \act]\phi)}
- \Pr(\intension{\K}{[\ACT = \act']\phi})),$$
so that $\delta_{\act,\act',\phi}$ measures how much more likely it is that
$\phi$ will result from performing $\act$ than from performing $\act'$
(except that if performing $\act'$ is more likely to result in $\phi$
than performing $\act$, we just take $\delta_{\act,\act',0}$ to be 0).

\commentout{
Let
$$\begin{array}{ll}
  \K_{\act,\act',\phi} = \\
    \intension{\K}{[\ACT \gets \act]\phi \land
  [\ACT \gets \act']\neg \phi} \union \intension{\K}{[\ACT \gets \act']\phi \land
            [\ACT \gets \act]\neg \phi}.\end{array}$$ 
That is, $\K_{\act,\act',\phi}$ consists
of all causal settings where setting $\ACT$ to $\act$ or $\act'$ affects
whether $\phi$ occurs.
In all the causal settings $(M,\vec{u}) \in
\K_{\act,\act',\phi} \inter \intension{\K}{[\ACT \gets \act]\phi}$,
$\ACT = \act$ is 
a \emph{but-for cause} of $\phi$: just changing the value of $\ACT$ (to
$\act'$) is enough to change the outcome; 
similarly, $\ACT = \act$ is a but-for cause of $\neg \phi$
in the causal settings in $\K_{\act,\act',\phi} \inter
\intension{K}{[\ACT \gets \act]\neg \phi}$.
\mtodo{is this definition doing the right thing when outcomes are
  overdetermined? If both Billy and Suzy have a switch available to
  switch the train onto the side track then and if either flips the
  switch the train goes down the side - don't both get 0 moral
  responsibility for all of the outcomes? since its only over but-for
  causes does it matter who flips the switch first? its true its not
  preventable...}
}
\commentout{
Let
$$\begin{array}{lll}
  \commentout{
  \delta_{\act,\act',\phi} = \\ \sum_{(M,\vec{u}) \in
         \K_{\act,\act',\phi}} \Pr(M,\vec{u}) 
      (\util(w_{M,\ACT \gets \act',\vec{u}}) 
         -  \util(w_{M,\ACT \gets \act,\vec{u}})).\end{array}$$ 
Thus, $\delta_{\act,\act',\phi}$ is the 
           difference between the expected utility of performing $\act$ and
$\act'$ in those worlds where which of these actions is performed
affects whether $\phi$ holds.
The actions $\act$ and
     $\act'$ might also have different expected utilities in worlds in
     $\K-\K_{\act,\act',\phi}$.
     However, we do not consider these worlds,
     since the choice of $\act$ vs.~$\act'$ has no impact on $\phi$
     (although it may have an impact on other variables that affect
     the utility).
     }
    \delta_{\act,\act',\phi} &= &\Pr(\intension{\K}{[\ACT=\act]\phi \land
    [\ACT=\act']\neg \phi}) -\\  &&\ \ \Pr(\intension{\K}{[\ACT=\act']\phi \land
    [\ACT=\act]\neg \phi}).\end{array}$$
Thus, $\delta_{\act,\act',\phi}$ is the 
           difference between the probability that performing $\act$
           will result in $\phi$ while performing $\act'$ won't, and
           the probability that performing $\act'$ will result in
           $\phi$ while performing $\act'$ won't. 
The difference $\delta_{\act,\act',\phi}$ can be viewed as a measure of how much
more like $\act$ makes $\phi$ than $\act'$.
}

\commentout{
\dfn\label{dfn:moralresp}
The \emph{effect of $\act$ on
$\phi$ given epistemic state $\E = (\Pr,\K,\util)$}, denoted
$\mr(\act,\phi,\E)$, is 
$\max_{\act'}\delta_{\act,\act',\phi}$.
\edfn

\mtodo{While most of the examples deal with negative moral
  responsibility we also want to capture positive moral
  responsibility? Is this captured with a different sign of $\delta$
  (with regard to societies utilities)? What about the case where the
  agent saves 1 in order to kill the two people that would have had a
  claim to his inheritance? This definition gives the agent moral
  praise for saving the 1... seems weird but I can see what its
  getting at. }

Since $\delta_{\act,\act',\phi}$ is a difference of probabilities, we
must have $-1 \le \delta_{\act,\act',\phi} \le 1$.
It is easy to see that $\delta_{\act,\act,\phi} = 0$, so
$\mr(\act,\phi,\E) \in [0,1]$.  
It is easy to check that $\mr(\act,\phi,\E) = 1$
if $\act$ guarantees $\phi$ and there is another action
$\act'$ that guarantees $\neg \phi$;
$\mr(\act,\phi,\E) = 0$ if the probability of $\phi$ with $\act$ is at
least as high as it is with any other action.
}

The difference $\delta_{\act,\act,\phi'}$ is clearly an important
component of measuring the blameworthiness of $\act$ relative to $\act'$.
But there is another component, which
we can think of as the cost of doing $\act$.  Suppose that Bob could
have given up his life to save Tom.  Bob decided to do nothing, so Tom died.
The difference between the probability of Tom dying if Bob does 
nothing and if Bob gives up his life is 1 (the maximum possible), but we do
not typically blame Bob for not giving up his life. 
What this points out is that blame is also concerned with the
\emph{cost} of an action.  The cost might be cognitive effort,
time required to perform the action, emotional cost, or (as in
the example above) death.  

We assume that the cost is captured by some outcome variables.
The cost of an action $\act$ is then the impact of performing $\act$ on
these variables.
We call the variables that we consider the \emph{action-cost} variables.
Intuitively, these are variables that talk about features of an action:
Is the action difficult? Is it dangerous?  Does it involve emotional upheaval?
Roughly speaking, the cost of an action is then measured by the
(negative) utility of the change in the values of these variables due
to the action.  There are two problems in making this precise: first,
we do not assign utilities to individual 
variables, but to worlds, which are complete settings of variables.  
Second, which variables count as
action-cost variables depends in part on the modeler.  That said, we do assume
that the action-cost variables satisfy some minimal properties.
To make these properties precise, we need some definitions.

Given a causal setting  $(M,\vec{u})$ and endogenous variables $\vec{X}$ in
$M$, let 
$w_{M,\vec{u}}$ denote the unique world determined by the causal setting
$(M,\vec{u})$ and let 
$w_{M, \vec{X} \gets \vec{x},\vec{u}}$ denote the unique world determined
by setting $\vec{X}$ to $\vec{x}$ in $(M,\vec{u})$.  Thus, for each endogenous variable
   $V$, the value of $V$ in world $w_{M, \vec{X} \gets \vec{x},\vec{u}}$ is
     $v$ iff $(M,\vec{u}) \sat [\vec{X} \gets \vec{x}](V=v)$.
Given an action $\act$ and outcome
variables $\vec{O}$, let $\vec{o}_{M,\ACT \gets \act, \vec{u}}$ be the
value of $\vec{o}$
when we set $\ACT$ to $\act$ in the setting $(M,\vec{u})$; that is,
$(M,\vec{u}) \sat [\ACT \gets \act](\vec{O} = \vec{o}_{M,\ACT \gets
  \act, \vec{u}})$.  Thus, $w_{M,\vec{O} \gets \vec{o}_{M,\ACT \gets
    \act, \vec{u}},\vec{u}}$ is the world that results when $\vec{O}$ is set
to the value that it would have if action $\act$ is performed in
causal setting $(M,\vec{u})$.  
To simplify the notation, we omit the $M$ and
$\vec{u}$ in the subscript of $\vec{o}$ (since they already 
appear in the subscript of $w$), and just write
$w_{M,\vec{O} \gets \vec{o}_{\ACT \gets   \act}, \vec{u}}$.
The world $w_{M,\vec{O} \gets \vec{o}_{\ACT \gets
    \act}, \vec{u}}$ isolates the effects of $\act$ on the variables
in $\vec{O}$.

With this background, we can state the properties that we expect
the set $\vec{O}_c$ of action-cost variables to have:
\begin{itemize}
\item for all causal settings $(M,\vec{u})$ and all actions $\act$, we have
  $$\util(w_{M, \vec{u}}) \ge \util(w_{M,\vec{O}_c \gets
    \vec{o}_{\ACT \gets   \act}, \vec{u}})$$
  (so performing $\act$ is actually costly, as far as the variables in
  $\vec{O}_c$ go);
\item for all causal settings $(M,\vec{u})$, all actions $\act$, and
  all subsets $\vec{O}'$ of $\vec{O}_c$, we have 
$$\util(w_{M,\vec{O}_c \gets \vec{o}_{\ACT \gets   \act}, \vec{u}}) \le
\util(w_{M,\vec{O}' \gets \vec{o}'_{\ACT \gets   \act}, \vec{u}})$$
(so all variables in $\vec{O}_c$ are costly---by not considering some
of them, the cost is lowered).
\end{itemize}

\dfn The \emph{(expected) cost} of action $\act$ (with respect to $\vec{O}_c$),
denoted $\cost(\act)$, 
is $\sum_{(M,\vec{u}) \in \K} \Pr(M,\vec{u}) (\util(w_{M,\vec{u}} -
  \util(w_{M,\vec{O}_c \gets \vec{o}_{\ACT \gets   \act},
    \vec{u}}))$. 
\edfn

If we think of $\phi$ as a bad outcome of performing $\act$, then the
blameworthiness of $\act$ for $\phi$ relative to $\act'$
is a combination of the likelihood to
which the outcome could have been improved by performing $\act'$ and
cost of $\act$ relative to the cost of $\act'$.  Thus, if $\cost(\act)
= \cost(\act')$,
the blameworthiness of $\act$ for $\phi$ relative to $\act'$ is just
$\delta_{\act,\act',\phi}$.   But if performing $\act'$ is quite costly
relative to performing $\act$, this should lead to a decrease in
blameworthiness.  How much of a decrease is somewhat subjective.  To
capture this, choose $\MX > \max_{\act'} \cost(\act')$ (in general, we
expect $\MX$ to be situation-dependent).
The size of  $\MX$ is a
measure of how important we judge cost to be in determining
blameworthiness; the larger $\MX$ is, the less we weight the cost.

\dfn\label{dfn:moralresp} The \emph{degree of blameworthiness of $\act$ for
$\phi$ relative to $\act'$ (given $\cost$ and $\MX$)}, denoted
$\mr_{\MX}(\act,\act',\phi)$, is 
$\delta_{\act,\act',\phi}\frac{\MX - \max(\cost(\act') - \cost(\act),0)}{\MX}$.
The degree of blameworthiness of $\act$ for $\phi$, denoted
$\mr_{\MX}(\act,\phi)$ is $\max_{\act'}\mr_{\MX}(\act,\act',\phi)$.
\edfn
Intuitively, we view the cost as a mitigating factor when computing
the degree of blameworthiness of $\act$ for $\phi$.   We can think of 
$\frac{\MX - \max(\cost(\act') -   \cost(\act),0)}{\MX}$ as the mitigation
factor.   No mitigation is
needed when comparing $\act$ to $\act'$ if
the cost of $\act$ is greater than that of $\act'$.
And, indeed, because of the  $\max(\cost(\act) -
\cost(\act'),0)$ term, if $\cost(\act) \ge
\cost(\act')$ then the mitigation factor is 1,
and $\mr(\act,\act',\phi) = \delta_{\act,\act',\phi}$.
In general, $\frac{\MX +
  \max(\cost(\act) -   \cost(\act'),0)}{\MX} \le 1$. 
  Moreover, $\lim_{\MX \rightarrow \infty} \mr_{\MX}(\act,\act',\phi) =
  \delta_{\act,act',\phi}$.  Thus, for large values of
  $\MX$, we essentially ignore the costliness of the act.  On the other
    hand, if $\MX$ and $\cost(\act')$ are both close to $\max_{\act''}
  \cost(\act'')$ and $\cost(\act) = 0$, then $\mr_{\MX}(\act,\act',\phi)$
is close to 0.  Thus, in the example with Bob and Tom above, if we
take costs seriously, then we would not find Bob particularly
blameworthy for Tom's death if the only way to save  Tom is for
Bob to give up his own life.

The need to consider alternatives when determining
blameworthiness is certainly not new, as can
be seen from the essays in \cite{WM03}.%
\fullv{\footnote{In the philosophy literature,
part of the discussion of alternatives is bound 
up with issues of determinism and free will (if the world is deterministic and
agents do not have free will, then they never could have done
otherwise).  In this paper, we ignore this issue, and implicitly assume
that agents always have choices.}}
What seems to be new here is the emphasis on blameworthiness with respect
an outcome and taking cost into account. 
The following example shows the impact of the former point.
\xam\label{xam:trolley1}
Suppose that agent $\ag$ is faced with the following dilemma:
if $\ag$ doesn't pull the lever, six anonymous people die; if $\ag$ does
pull the lever, the first five people will still die, but the sixth will
be killed with only probability 0.2.
%
If $\ag$ does not pull the lever, $\ag$ is not blameworthy for
the five deaths (no matter what he did, the five people would have
died), but has some degree of blameworthiness for the sixth. The point here is
that just the existence of a better action $\act'$ is not enough.
To affect $\ag$'s blameworthiness for outcome $\phi$, 
action $\act'$ must be better in a way that affects $\phi$.
\exam

\commentout{
\noindent {\bf Example~\ref{xam:trolley1} (cont'd):}
Suppose that the agent assigns utility 1 to no one dying, utility 0
to 5 people dying and there being significant property damage, and
utility .1 to 5 people dying with no property damage. Let $O_1 = 1$ if
5 people die and $O_1 = 0$ if no one dies; let $O_2 = 1$ if there is
significant property damage and $O_2 = 0$ if there is no property damage.
If the only 
possible actions are $\act$, which results in significant property
damage and 5 people dying, and $\act'$, which results in 5 people dying
and significant property damage with probability .2 and in 5 people
dying and no property damage with probability .8, then the
blameworthiness of $\act$ for $O_1 = 1$ is 0, since $\K_{\act,\act',O_1
= 1} = \emptyset$; there are no worlds where $\act$ and $\act'$
differ insofar as the number of people who die.  However,
the moral responsibility of $\act$ for $O_1 = 1 \land O_2 = 1$ (or
just for $O_2 = 1$) is $.8 \times .1 = .08$.   Of course, this depends
very much on the utility function; there will be many who find it
repugnant that property damage can affect utility so much in light of
5 people dying.  We return to this issue in
Section~\ref{sec:conclusion}. \bbox
}

\mtodo{what follows doesn't quite match the definition since it is
  describing the blameworthiness/praiseworthiness of the action not
  the action for a specific outcome. }
\commentout{
Moral responsibility as defined here is a measure of
\emph{blameworthiness}; the higher the moral responsibility of the action
$\act$ that $\ag$ performs for the outcome $\phi$, the worse $\act$
is, at least as far as $\phi$ is concerned.  But suppose that we
want to praise $\ag$ for performing a ``good'' action that led to
$\phi$?  In that case,
we would consider the dual notion, the \emph{degree of praiseworthiness} of
$\act$ for $\phi$ given $\E$, denoted $\pw(\act,\phi,\E)$, to be
$\min_{\act' \ne \act} 
\delta_{\act',\act,\phi}$.  The degree of praiseworthiness
can be viewed as a measure of how much better $\act$ is for $\phi$
than the best of
the alternatives.
\mtodo{can we say something here that praiseworthiness is will be
  bounded by like we did for def 0.6? that will help people get the
  right intuition}
Note that the moral praiseworthiness of an action can range from $-1$
to $1$.  
\mtodo{These definition make an interesting prediction. Say there are 3 option, let a person die, save them with P=0.5 and save them with P=1. If the person chooses the middle option and the person is saved, they will be praised for outcome? Yet if the person dies they will be blamed for the outcome. Can we generalize this and say that failure to act fully optimally means that a person can be both blamed and praised depending on the outcome while this won't be true for an optimal action (its not blameworthy). }
}

\commentout{
As the next example shows, even if an agent chooses the action that 
maximizes expected utility, he may have a positive degree of blameworthiness
for a bad outcome.

\xam\label{xam:maxutility1}
Consider a variant of the trolley problem where the track looks as
shown in Figure~\ref{fig:dogs}
\begin{figure}
\begin{center}
\begin{picture}(6,4.4)(2,1.5)
\setlength{\unitlength}{.24in}
\put(2,4){\line(-1,-2){1}}
\put(6,4){\line(-1,-2){1}}
\put(2,4){\line(1,-2){1}}
\put(6,4){\line(1,-2){1}}
\put(4,6){\line(-1,-1){2}}
\put(4,6){\line(1,-1){2}}
\put(2.25,4.7){$C$}
\put(6.65,2.9){$P$}
\end{picture}
\end{center}
\caption{Colonel and Perdita on the tracks.}
\end{figure}

Cruella has abducted (dognapped?) two dogs, Colonel
and Perdita, and has tied them to track 1 and track 4, respectively.
Roger does not know whether the train will go left
or right at the first junction.  (Formally, this is characterized by a
variable $\LT$ which is 1 if the train goes left and 0 if the train
goes right; $\LT$ is determined by the exogenous variable.) He
ascribes probability $1/2$ to each possibility.
He can choose to flip a switch left
or right, represented by the acts $\act_L$ and $\act_R$; this 
determines whether the train goes left or right at
the second junction.  But he must make his decision immediately
(before knowing whether the train goes left or right at the first
junction) for his choice to be effective. As the figure shows, 
if the train goes left and he chooses $\act_L$, then Colonel dies; 
if the train goes right and he chooses $\act_R$, then Perdita dies;
otherwise, neither dies.
There are four worlds, which can be succinctly described
by the values of $\LT$, $\CL$ (Colonel lives), and $\PL$ (Perdita
lives): $w_{LL} = (\LT=1,\CL=0,\PL=1)$, $w_{LR} = (|LT=1,\CL=1,\PL=1)$,
$w_{RL} = (\LT=0,\CL=1,\PL=1)$, $w_{RR} = (\LT=0,\CL=1,\PL=0)$.
If $\LT=0$, then $\CL=1$ no matter what Roger does; what Roger does
matters only if $\LT=1$.  If Roger chooses $\act_L$, then
the outcome is either $w_{LL}$ or $w_{RL}$, which each occur with probability
$1/2$, so Colonel
does with probability $1/2$ and Perdita definitely does not die.  If
Roger chooses $\act_R$, then the outcome is either $w_{LR}$ or $w_{RR}$,
which each occurs with probability $1/2$, so 
Perdita dies with probability $1/2$ and Colonel definitely does not die.
Since Colonel is old, if one of them must die, Roger marginally
prefers that it be Colonel.  For definiteness, suppose that
$\util(w_{LL}) = 0$, $\util(w_{LR}) = \util(w_{RL}) = 1$, and $\util(w_{RR}) = .1$.
Thus, Roger maximizes expected utility by
choosing $\act_L$.  Unfortunately, the train goes left, and Colonel
dies.  Note $\K_{\act_L,\act_R, \CL=0} = \{w_{RL},w_{RR}\}$, so Roger's
degree of blameworthiness for $\CL=0$ is  $.45$, despite choosing the
action that maximized expected utility.
Moreover, the degree of preventability of $\CL=0$ is 1; if Roger had
flipped the switch right rather than left, Colonel's death would
definitely have been prevented.
\exam
}

Defining the degree of blameworthiness of an
action for a particular outcome, as done here, seems to be consistent
with the legal view.  A prosecutor considering what to
charge a defendant with is typically considering which outcomes that
defendant is blameworthy for.

\commentout{
However, there is also a large literature on the
\emph{moral permissibility} of an action $\act$ that does not mention a specific
outcome.  We can identify the moral permissibility of an
action $\act$ with the degree of blameworthiness of the agent for $\ACT \gets
\act$.  Intuitively, this is because the computation of the
blameworthiness of $\ACT=\act$ compares the utility of performing
$\act$ to the utility of performing $\act'$.

\pro If $\act$ is the action that maximizes expected utility with respect
to $\E$, then the degree of blameworthiness of $\act$ for the outcome
$\ACT=\act$ is 0.
\epro

The key observation here is that if $\act \ne \act'$, then
$\K_{\act,\act',\ACT = \act} = \K$.  Thus, we consider all possible causal
settings when evaluating an agent's degree of blameworthiness
for $\ACT = \act$.  In the basic trolley example, the agent who chose
to pull the lever is blameworthy for the death of one person.
Most people would feel badly about the outcome, but still feel that it
was the right thing to do.
Intuitively, by considering the
blameworthiness of $\act$ for the outcome $\ACT = \act$, we are
considering the ``bigger picture'', by considering all of $\K$ rather
than just some subset of it.
}

Blameworthiness is defined relative to
a probability distribution.
We do not necessarily want to use the agent's subjective
probability.  For example, suppose that the agent had several
bottles of beer, goes for a drive, and runs over a pedestrian.  
The agent may well have believed that the probability 
that his driving would cause an accident was low, but we clearly don't
want to use his subjective probability that he will cause an accident
in determining blameworthiness. 
Similarly,
suppose that a doctor honestly believes that
a certain medication will have no harmful side effects for a patient.
One of his patients who had a heart condition takes the medication and
dies as a result.  If the literature distributed to the doctor
included specific warning about dire side-effects for patients with this
heart condition but the doctor was lazy and didn't read it, again, it
does not seem reasonable to use the 
doctor's probability distribution.
Rather, we want to use the probability distribution that he should
have had, had he read the relevant literature.
\fullv{Our definition allows
us to plug in whatever probability distribution we consider most appropriate.}
\shortv{We can use whatever probability distribution we consider
  most appropriate in the definition.}

\commentout{
The same type of issue arises when considering utility.  We don't want
the fact that a drug lord wants to kill his rival to be a reason not to
hold him blameworthy for the death.    
In cases like this, the courts typically consider the
probability and utility that a ``reasonable person'' would have,
rather than that of the agent.  For ascribing moral responsibility, this
seems reasonable.%
\fullv{\footnote{Note that considering a reasonable person's
    utility does not mean that we must use the same utility function
    for all agents.  For example, in the trolley problem, a reasonable
    person might well ascribe 
    higher utility to saving, say, his wife than to saving a random
    person. It is also worth noting that
one situation where we do not want to use a ``reasonable person's''
probability is in the case of diminished responsibility, such as
insanity or if the agent is a 5-year-old.  The so-called
\emph{M'Naghten test} \cite{Elliott96} 
essentially says that an agent should not be held
morally responsible in the eyes of the law if he did not know ``the
nature and quality of the action he was doing''.  Certainly one part of
knowing the nature and quality of an action involves having a reasonably accurate
estimate of the probability that it will cause a particular outcome
and having an understanding of its consequences so as to be able to
judge utility.
Thus, if there is independent reason to believe that an agent $\ag$ is
sufficiently impaired so that the M'Naghten test applies, then $\ag$
should not be held morally responsible; specifically, we should not
use a ``reasonable person's'' utility and probability in determining
blameworthiness.}} 
}

\mtodo{I think this is only partially
  true... do we think insane people just don't know the estimates of
  different probabilities or rather do we think that insane people
  fail in a computational sense rather than an epistemic sense of
  failing to act as utility maximizing rational agents?}\jtodo{I
  suspect it's all of the above.  Typically I don't think of the
  problem with insane people as being computational.  Rather, they
  simply have different probabilities and/or utilities than the rest
  of society, but we don't blame them for it.} 


In using the term ``blameworthiness'', we have implicitly been thinking
of $\phi$ as a bad outcome.  
If $\phi$ is a good outcome, 
it seems more reasonable to use the term
``praiseworthiness''.
However, defining praiseworthiness raises some significant new issues.
We mention a few of them here:
\begin{itemize}
\item Suppose that all actions are costless, Bob does nothing and,
  as a result, Tom lives.
Bob could have shot Tom, 
so according to the definition Bob's degree of blameworthiness
for Tom living is 1.  Since living is a good outcome, we may want to
talk about praiseworthiness rather than blameworthiness, but it still
seems strange to praise Tom for doing the obvious thing.
This suggests that for praiseworthiness, we should compare the action
to the ``standard'' or ``expected'' thing to do.
To deal with this, we assume that there is a \emph{default action} $a_0$,
which we can 
typically think of as ``doing nothing'' (as in the example
above), but does not have to be.
Similarly, we typically assume that the default action has low cost, but
we do not require this.
The praiseworthiness of an act is typically compared just to the
default action, rather than to all actions.
Thus, we  consider just $\delta_{\act,\act_0,\phi}$, not
$\delta_{\act,\act',\phi}$ for arbitrary $\act'$.
\item
  It does not seem that there should be
a lessening of praise if the cost of $\act$ is even lower than that of
the \fullv{default (although that is unlikely in practice).}
\shortv{default.}
On the other hand, it
seems that there should be an increase in praise the more costly
$\act$ is. For example, we view an action as particularly praiseworthy
if someone is 
risking his life to perform it.
This suggests that the degree of praiseworthiness of $\act$ should be
$\delta_{\act,\act_0,\phi}$ if $\cost(\act) \le
\cost (\act_0)$, and $\delta_{\act,\act_0,\phi}\frac{M - \cost(\act_0)
  + \cost(\act)}{M}$ if $\cost(\act) > \cost(\act_0)$.  But this has
the problem that the degree of praiseworthiness might be greater than
1.
\fullv{
To deal with this, we take the degree of praiseworthiness for
$\phi$ to be $\delta_{\act,\act_0,\phi}+
(1-\delta_{\act,\act_0,\phi})\frac{M - \cost(\act_0) +
  \cost(\act)}{M}$ if $\cost(\act) > \cost(\act_0)$.
(Some other function that increases
to 1 the larger $\cost(\act)$ is relative to $\cost(\act_0)$ would
also work.)}
\fullv{

But there is an additional subtlety.}
If  the agent put a lot of effort into $\act$ (i.e.,
  $\cost(\act) - \cost(\act_0)$ is large) because his main focus was
  some other outcome $\phi' \ne \phi$, and there is another
  action $\act'$ that would achieve $\phi$ at much lower cost, then
  it seems unreasonable to give the agent quite so much praise for his
  efforts in   achieving $\phi$.  \fullv{We might want to consider the effort
  required by the least effortful action that achieves $\phi$.}
\item We typically do not praise
  someone for an outcome that was not
intended (although we
  might well find someone blameworthy for an unintended outcome).
\end{itemize}
\shortv{In the full paper, we give a definition of praiseworthiness that takes
these concerns into account.}

\fullv{Putting all these considerations together, we have the
  following definition of praiseworthiness.
\dfn The \emph{degree of praiseworthiness of $\act$ (relative to $M$) for
  $\phi$}, denoted $\pw_M(\act,\phi)$, is 0 if $\phi$ was not an
intended outcome of $\act$ (as defined in the next section), and is 
is $\delta_{\act,\act_0,\phi} + \max(0,(1-\delta_{\act,\act_0,\phi})
\min_{\{\act': \delta(\act',\act_0,\phi) \ge \delta(\act,\act_0,\phi)\}}
\frac{M - \cost(\act_0) + \cost(\act')}{M})$ if $\phi$ is intended.
\edfn
This definition considers only acts $\act'$ that are at least as effective at
achieving $\phi$ as $\act$ (as measured by
$\delta_{\act',\act_0,\phi}$).  We could also consider the cost of acts that are
almost as effective at achieving $\phi$ as $\act$.   We hope to
do some experiments to see how people actually assign degrees of
praiseworthiness in such circumstances.
}


\commentout{
\dfn The \emph{degree of praiseworthiness of $\act$ (relative to $M$) for
  $\phi$}, denoted $\pw_M(\act,\phi)$ is
$\delta_{\act,\act_0,\phi}\frac{M - \cost(\act_0) + \cost(\act)}{M}$.
\edfn
Note that if $\cost(\act) > \cost(\act_0)$, then $\pw_M(\act,\phi) >
\delta_{\act,\act_o,\phi}$.  The more costly an action is, the more
praiseworthy it is.
}


The focus of these definitions has been on the blame (or praise) due
to a single individual.  Things get more complicated once we consider groups.
Consider how these definitions play out in the
context of the \fullv{well-known} \emph{Tragedy of the Commons}
\cite{Hardin68}, where there are many agents, each of which can perform
an action (like fishing, or letting his sheep graze on the commons)
which increases his individual utility, but if all agents perform the
action, they all ultimately suffer (fish
stocks are depleted; the commons is overgrazed). 

\xam\label{xam:tragedy}
Consider a collective of fishermen.  Suppose that if
more than a couple of agents fail to limit their fishing, the fish stocks
will collapse 
and there will be no fishing allowed the
following year.
The fisherman in fact all do fish, so the fish stocks collapse.
\commentout{
We can assume that each agent has ``reasonable'' utilities: they
prefer that there be no collapse, but if there is going to be
overfishing anyway, they would prefer to fill up their boats with fish
this year. 
Suppose that in fact no agents limit their fishing,
so fishing is not
allowed the following year.
}

Each agent is clearly part of the cause of
the outcome.
To determine a
single agent's degree of blameworthiness, we must consider that 
agent's uncertainty about how many of the other fisherman will 
limit their fishing.  If the agent believes (perhaps justifiably)
that, with high probability,
very few of them will limit their fishing,
then his blameworthiness will be quite low.
As we would expect, under minimal assumptions about the probability
measure $\Pr$, the more fisherman there are and the larger the gap between the
expected number of fish taken and the
number that will result in overfishing limitations, the lower the
degree of blameworthiness.
Moreover, a fisherman who catches less fish has less blameworthiness.
In all these cases, it is less likely that changing 
his action will lead to a change in outcome.
\exam

The way that blameworthiness is assigned to an individual fisherman in
Example~\ref{xam:tragedy} essentially takes the actions of
all the other fisherman as given.  But it is somewhat disconcerting
that if each of $N$ fisherman justifiably believed that all the other
fisherman would overfish, then each might have degree of
blameworthiness significantly less than the $1/N$ that we might
intuitively give them if they all caught roughly the same number of fish. 

One way to deal with this is to consider the degree of blame we
would assign to all the fisherman, viewed as a collective (i.e., as a
single agent).
The collective can
clearly perform a different action that would
lead to the desired outcome.  Thus, viewed as a
collective, the fishermen have 
degree of blameworthiness close
\fullv{to 1 (since they could performed a
joint action that resulted in no further fishing, and they could have
performed an action that would have guaranteed that there would be
fishing in the future).}
\shortv{to 1.}

How should we  allocate this ``group moral blameworthiness'' to the
individual agents? 
We believe that Chockler and Halpern's \citeyear{ChocklerH03} notion of
responsibility and blame can be helpful in this regard, because they are
intended to measure how responsibility and blame are diffused in a
group.   It seems that when ascribing moral responsibility in group
settings, people consider both an agent as an individual and as a
member of a group.  Further research is needed to clarify this issue.

\commentout{
\subsection{Group Blameworthiness}
Our definition of moral blameworthiness and praiseworthiness captures
the extent that an individual could have made a difference to the
outcome if they were to consider society's utility and rational
beliefs about the world. [[Add or refer to example of person acting on
reasonable but inaccurate beliefs that cause a problem]]. However the
normativity of beliefs when other rational agents add another layer of
complexity. [[Add example agent believes nature will do X vs. an
example where agent believes person will do X give an intuition of why
we might not necessarily want to go that far with just belief]]. This
gives us a sense that to asses to amount of blameworthiness and
praiseworthiness for an agent acting among other agents, the amount of
moral responsibility we attribute may also depend on the action space
of the other agents.

\xam\label{xam:poison} Suppose that Louis asks Sibella to buy him some
rat poison at the store. It's possible that Louis has a rat problem he
is trying to deal with but it is also possible that he plans to use
the rat poison to kill Rufus. From the perspective of society, the
relative ranking of possible outcomes is: Rufus alive + rats dead >
Rufus alive + rats alive > Rufus dead + rats alive. Sibella chooses to
not buy the rat poison at the store.

First consider that Sibella believes that Louis plans to use the rat
poison to kill Rufus. In this case, Sibella refused to buy the rat
poison and Rufus lives but the rats also lived. If we use Sibella's
beliefs as the epistemic state for calculating his blameworthiness
then we should not find him blameworthy for keeping the rats alive and
would find him praiseworthy for Rufus staying alive.

Now consider that Sibella believes that Louis plans to use the rat
poison to resolve his rat problem. In this case, Sibella's refusal to
buy the rat poison make him blameworthy for the outcome of the rats
alive. Since the epistemic state is an input to the definitions, these
predictions are true whether or not Sibella's beliefs were actually
true. The point here is that we must consider the appropriate belief
state when calculating blameworthiness. However it is not always
obvious that the agent's belief state is the one we should consider.
\exam

\xam\label{xam:commons} There is a group of 1000 fisherman that all
fish from the same common pool of fish. If their total monthly take is
10,000 or less tons of fish the fish stock will recover the next
month. If they cumulatively exceed 10K tons the population will
collapse and the fish stock will take over a year to recover making
everyone worse off. Each fisherman can choose to fish up to 20 tons
each per month. The preferences of each fisherman are: less than
10,000 tons fished in total + I catch $k$ tons ($k \le 20$) > less than
10,000 tons fished in total + I catch $k'$ tons ($k' < k$) > more than
10,000 tons fished in total + I catch $k$ tons ($k \le 20$) > more than
10,000 tons fished in total + I catch $k'$ tons ($k' < k$). The
preferences of society are: less than 10,000 tons fished in total >
more than 10,000 tons fished in total.

Specifically consider two fisherman A and B, two of the 100 fishermen.
Now suppose that fisherman A believes that many other fisherman will
overfish, and the total catch will be well over 10,000 tons; the
probability of it being under 10,000 given he fishes 20 is $\rho_A$. A
thus catches 20 tons. To what degree is A blameworthy for the outcome?
How about fisherman B who catches only 9 but also believes that his
action will put the total over 10,000 with probability
$\rho_B = \rho_A$? Are they equal to blame? Does the relative
difference in their actions (9 vs. 20) or their beliefs affect their
degree of blameworthiness? \exam

\xam\label{xam:coastguard} There are two small boats about to sink.
Boat A has 10 people on board and boat B has 5 people on board. There
are two coastguards, Billy and Suzy who must choose which boat to
rescue. It requires two coastguards to jointly rescue the people on a
boat, one coastguards will have no effect. Billy and Suzy are the only
nearby coastguards and there is only time for them to choose one
boats. Their preferences (and the preferences of society) are: save 10
(A,A) > save 5 (B,B) > save 0 (A,B or B,A). Thus the game has two pure
strategy Nash equilibria (A,A) and (B,B).

However Billy believes that Suzy is going to rescue boat B with
probability 0.9 and likewise Suzy believes that Billy is going to
rescue boat B with probability 0.9. Both choose to rescue boat B and 5
people are saved. To what extent are they blameworthy for letting the
10 people die? More generally, when individuals have to coordinate in
order to realize a socially desirable outcome, to what extent will
they be considered blameworthy for a suboptimal outcome when they did
the best they could have considering their beliefs.

\exam

\xam\label{xam:shooting}

Assume there is a group of $N$ gang members
that have a plan to murder a rival gang member out of vengeance. Each
individual gang member would rather not murder their rivals also fear
that the gang leaders will reprimand those who are perceived as
cowardly and do not shoot. If none of them shoot they will not be
reprimanded. Thus the preferences of each gang member are: rival lives
(no one shoots) > rival dies (I'm one of the shooters) > rival dies
(I'm \emph{not} one of the shooters).

When the number of gang members is small, it is possible that they
will coordinate on a strategy where nobody shoots. However, as the
number of gang members grows it becomes less and less likely that not
a single one will shoot and thus it starts to be in everyone's best
interest to shoot. Our definitions make the prediction that since as
$N$ increases each gang members rational belief that the others will
act also grows. As discussed above the more a gang member believes
that the other agents will shoot the less likely he is to be a but-for
cause and hence the less likely to be held morally responsible. Thus
this feature of our definition gives another sense of the diffusion of
moral responsibility. \exam

Furthermore, in some situations agents may collectively have beliefs
that lead to rationally consistent outcomes. [[give the lifeboats
examples]]. This problem is often exacerbated when individual
incentives diverge from what society desires. [[give pollution or gang
shooting case or even a straight forward public good game]]. However
we still want agents to act based on their beliefs [[poison case from
scanlon?]]. To understand why we have an intuition that blame should
be assigned differentially we consider that a full account of
blameworthiness will also include group blame -- which allow us to
assign blame to agents who acted in a socially desirable way
conditional on their beliefs but were a part of a group which failed
to find a socially optimal equilibrium.
}

\section{Intention}\label{sec:intention}
The definition of degree of blameworthiness does not take
intention into account.
In the trolley problem, an agent who pulls the lever so that only one
person dies is fully blameworthy for that death.
However, it is clear that the agent's intent was to save five people,
not kill one; the death was an unintended side-effect.
Usually, agents are not held responsible for accidents and the moral
permissibility of an action does not take into account
unintended side-effects.

Two types of intention have been considered in the literature (see, e.g., \cite{CL90}):
(1) whether agent $\ag$ intended to perform action $\act$ (perhaps it was
an accident) and (2) did $\ag$ (when performing $\act$) intend outcome
$\phi$ (perhaps $\phi$ was an unintended side-effect of $\act$, which was
actually performed to bring about outcome $o'$).
Intuitively, an agent intended to perform $\act$ (i.e., $\act$ was not
accidental) if his expected 
utility from $\act$ is at least as high as his expected utility from
other actions.
The following definition formalizes this intuition.

\dfn\label{dfn:accident}
Action $\act$ was \emph{intended in
  $(M,\vec{u})$ given epistemic state $\E = (\Pr,\K,\util)$} if
$(M,\vec{u}) \sat \ACT =
\act$ ($\act$ was actually 
performed in  causal setting $(M,\vec{u})$),
$|\R(\ACT)| \ge 2$ ($\act$ is not the only possible action), and for all 
$\act' \in \R(\ACT)$,
     $$\sum_{(M,\vec{u}) \in \K} \Pr(M,\vec{u}) 
(\util(w_{M,A \gets \act,\vec{u}}) - \util(w_{M,A \gets \act', \vec{u}})) \ge 0.$$
\edfn


The assumption that $|\R(\ACT)| \ge 2$ captures the intuition that we
do not say that $\act$ was intended if $\act$ was the only action that
the agent could perform.   We would not say that someone who is an epileptic
intended to have a seizure,
since they could not have done otherwise. 
What about someone who performed an action because there was a gun
held to his head?  In this case, it depends on how we model the set
$\ACT$ of possible actions.  If we take the only feasible action to be
the act $\act$ that was performed (so we view the agent as having no real
choice in the matter), then the action was not intended.  But if
we allow for the possibility of the agent
choosing whether or not to sacrifice his life,
then we would view 
whatever was imposed as intended.  

\fullv{
Requiring that $|\R(\ACT)| \ge 2$  also lets us deal with some
standard examples in the philosophy literature.
For example,  Davidson \citeyear{Davidson80} considers
  a climber who knows that he can save himself from plummeting to
  his death by letting go of a rope connecting him to a companion who
  has lost his footing, but the thought of the contemplated action so
  upsets him that he lets go accidentally (and hence unintentionally).
  We would argue that at the point that the climber let go of the rope,
  he had no alternative choices, so the action was not intended, even
  if, had he not gotten upset, he would have performed the same
action at the same time intentionally (because he would then have had
other options).
}

  \commentout{
  Our   definition can also be viewed as capturing ``revealed intention'',
  in the same sense that decision theorists talk about ``revealed
  preference''; that is, an agent can say that he intends to stop
  smoking, but yet does not stop.   So (at least according to the
  standard decision-theoretic argument) his utilities must be such
    that he actually prefers smoking to not smoking, despite his claim.
  These examples show some of the subtleties involved in capturing a
  notion of intention;  that said, we believe that our definition covers
  most standard usages of the word ``unintenional''.
}
  
\commentout{
  It is also possible to define the circumstances under which the
outcome of an action is intended. Our approach is based on the
definitions of intention in Kleiman-Weiner et \shortv{al.~\citeyear{KWG15},} 
\fullv{al.~\citeyear{KWG15,KWGprep},} and gets
essentially the same  
results. However, the work 
presented here directly leverages the
structural-equations framework.
}
The intuition
for the agent intending outcome $\vec{O} = \vec{o}$ is that, 
had $\act$ been unable to affect $\vec{O}$, $\ag$ would
not 
have performed $\act$.
But this is not quite right for several reasons, as the following
examples show.


\xam\label{xam:doctor}  Suppose that a patient has malignant
lung cancer.  The only thing that the doctor believes that he can do to
save the patient is to remove part of the lung.  But this operation is
dangerous and may lead to the patient's death.  In fact, the patient does die.
Certainly the doctor's operation is the cause of death, and
the doctor intended to perform the operation.  However, if the variable $O$
represents the possible outcomes of the operation, with $O=0$ denoting
that the patient dies and $O=1$ denoting that the patient is cured,
while the doctor intended to affect the variable $O$, he certainly did
not intend the actual outcome $O=0$. \exam
\xam\label{xam:louisintent} Suppose that Louis plants a bomb at a
table where his cousin Rufus, who is 
  standing in the way of him getting an inheritance, is going to have
  lunch with Sibella.
Louis get 100 units of utility if Rufus dies, 0 if he doesn't die,
and $-200$ units if he goes to jail.  His total utility is the sum of
the utilities of the relevant \shortv{outcomes.} \fullv{outcomes (so,
  for example, $-100$ if Rufus dies and he goes to jail).} 
He would not have planted the bomb
if doing so would not have affected whether Rufus dies.  On the other
hand, Louis would still have planted the bomb even if doing so had no
impact on Sibella.  Thus, we can conclude that Louis
intended to kill Rufus but 
did not intend to
kill Sibella.
  
    
     Now suppose that Louis has a different utility function, and prefers
     that both Rufus and Sibella die. Specifically,
  Louis get 50 units of utility if Louis dies and 50 units of utility if
     Sibella dies.  Again, he
gets $-200$ if he goes to jail, and his total utility is the sum of
the utilities of the relevant outcomes.
With these
utilities, intuitively, 
    Louis intends both Rufus and Sibella to die.
    Even if he
        knew that planting the bomb had no impact on whether Rufus lives
   (perhaps because Rufus will die of a heart attack, or because Rufus
   is wearing a bomb-proof vest), Louis would still plant the bomb 
    (since he would get significant
    utility from Sibella dying).
    Similarly, he would plant the bomb even if it had no
    impact on Sibella. 
Thus, according to the naive definition, Louis did not intend to kill
either Rufus or Sibella.        
\exam


\commentout{
  To deal with the second problem, say that $\vec{O}' = \vec{o}'$ is an
\emph{extension} of $\vec{O} = \vec{o}$ if every conjunct of $\vec{O}
= \vec{o}$ is a conjunct of $\vec{O}' = \vec{o}'$.
In order to show that $\ag$ intended
$\vec{O} = \vec{o}$ by performing $\act$, we require only that there is
some extension $\vec{O}' = \vec{o}'$ such that
$\ag$ intended every conjunct of $\vec{O}' = \vec{o}'$, in the sense
that $\ag$'s utility depends on all the variables in $\vec{O}'$, and
$\vec{O}'$ is minimal in this regard.
}

\commentout{
There is one more technical issue:
To ensure that an agent does not intend an outcome
like $\ACT = \act'$, we
assume that an agent's utility is 
not affected by the value of $\ACT$.  If two worlds agree on all but
the value of $\ACT$, then their utility must be the same.
This may at
first seem unreasonable.
For example, suppose that James is deciding
whether to drive a neighbor's unattended Mustang.  We can
imagine that just 
driving the Mustang would give James positive utility.  We can deal
with this by taking the features of driving that give James
positive utility (being in control of a powerful car, the sound of the
revving engine, and so on) to be separate variables whose value depends
on whether James drives the Mustang, and have James' utility depend
on these variables, rather than on the action itself.
Given these assumptions, we can now define what it 
means for an agent to intend an outcome.
}

Our definition will deal with both of these problems.
We actually give our definition of intent in two steps.  First, we
define what it means for agent $\ag$ to intend to affect the variables in
$\vec{O}$ by performing action $\act$.
\commentout{
Roughly speaking, the set of variables that $\ag$ intends to affect by
performing $\act$
is the complement of the set of costly variables.  This is not quite
right for two reasons. First, costly variables are not defined
relative to a particular action.  Second, we do not want to consider
variables whose value is unaffected by actions; if, no matter what
$\ag$ does, variable $O$ has a value that gives utility 100, we do not
want to say that $\ag$ intends to affect $O$.  We deal with these
problems by again assuming that we have a default action $\act_0$, and
comparing the effect of $\act$ to that of $\act_0$.  We think of
$\act_0$ as a low-cost action compared to $\act$.  The idea is that
agent $\ag$ intends to affect the variables in $\vec{O}$ if he would prefer
to perform $\act_0$ to performing $\act$ if the  
distribution of outcomes $\vec{O}$ was fixed at what he believes it
would be if he performed  $\act$.}

To understand the way we formalize this intuition better, suppose
first that $\act$ is deterministic.  Then $w_{M,\ACT
  \gets \act, \vec{u}}$ is the world that results when action $\act$
is performed in the causal setting $(M,\vec{u})$
and $w_{M,(\ACT\gets
    \act',\vec{O} \gets \vec{o}_{\ACT \gets \act}),\vec{u}}$ is the
world that results when act $\act'$ is performed, except that
the variables in $\vec{O}$ are set to the values that they would have had
if $\act$ were performed rather than $\act'$.  
 If $\util(w_{M,\ACT \gets \act, \vec{u}}) < \util(w_{M,(\ACT\gets
    \act',\vec{O} \gets \vec{o}_{\ACT \gets \act}),\vec{u}})$,  
that means that if the
  variables in $\vec{O}$ are fixed to have the values they would have
  if $\act$ were performed, then the agent would prefer to do
  $\act'$ rather than $\act$.
  Similarly, 
$\util(w_{M,(\vec{O} \gets \vec{o}_{\ACT \gets \act}),\vec{u}}) > 
\util(w_{M,(\vec{O} \gets \vec{o}_{\ACT \gets \act'}),\vec{u}})$ says
that the agent prefers how $\act$ affects the variables in $\vec{O}$
to how $\act'$ affects these variables.  
Intuitively, it will be these two conditions that suggest that the agent
  intends to affect the values of the variables in $\vec{O}$ by
  performing $\act$; once their values are set, the agent
  would prefer $\act'$ to $\act$.

  The actual definition of the agent intending to affect the variables
  in $\vec{O}$ is slightly more complicated than this in several
  respects.  First, if the outcome of $\act$ is probabilistic, we need
  to consider each of the possible outcomes of performing $\act$ and
  weight them by their probability of occurrence.  To do this, for
  each causal setting $(M,\vec{u})$ that the agent considers possible,
  we consider the effect of performing $\act$ in $(M,\vec{u})$ and
  weight it by the probability that the agent assigns to
  $(M,\vec{u})$. Second, we must deal with the situation discussed in 
Example~\ref{xam:louisintent} where Louis intends both Rufus and
Sibella to die.  Let $D_R$ and $D_S$ be variables describing whether
Rufus and Sibella, respectively, die.  While Louis certainly
intends to affect $D_R$, he 
will not plant the bomb only if both Rufus and
Sibella die without the bomb (i.e., only if both $D_R$ and $D_S$ are set to 0).
Thus, to show that the agent intends to affect the variable $D_R$, we
must consider a superset of $D_R$ (namely, $\{D_R,D_S\}$).  Third,
we need a minimality condition.  If Louis intended to kill only Rufus,
and Sibella dying was an unfortunate byproduct, we do not want to say
that he intended to affect $\{D_R,D_S\}$, although he would not have
planted the bomb if both $D_R$ and $D_S$ were set to 0.
There is a final subtlety: when considering whether $\ag$ intended to
perform $\act$, what alternative actions should we compare $\act$ to?
The obvious answer is ``all other actions in $\ACT$''.  Indeed, this
is exactly what was done by Kleiman-Weiner et al.~\citeyear{KWGprep}
(who use an approach otherwise similar in spirit to the one proposed
here, but based on influence diagrams rather than causal models).
We instead generalize to allow a \emph{reference set
$\REF(\act)$} of actions that does not include $\act$ but, as the
notation suggests, can depend 
  on $\act$, and compare $\act$ only to actions in $\REF(\act)$.
As we shall see, we need this generalization to avoid some problems.  
We discuss $\REF(\act)$ in more detail
\fullv{below, after giving the definition.} 
\shortv{below.}

    \dfn\label{dfn:intention} An agent $\ag$ \emph{intends to affect
            $\vec{O}$  by doing action $\act$ given epistemic state $\E =
      (\Pr,\K,\util)$
            and reference set $\REF(\act) \subset \R(\ACT)$}
    if and only if there exists a superset $\vec{O}'$
of $\vec{O}$  
such that (a)
$\sum_{(M,\vec{u}) \in \K} \Pr(M,\vec{u})
\util(w_{M,\ACT \gets   \act, \vec{u}}) \le
\displaystyle{\max_{\act' \in \REF(\act)}} 
\displaystyle{\sum_{(M,\vec{u}) \in \K}} \Pr(M,\vec{u}) 
\util(w_{M,(\ACT \gets \act', \vec{O}' \gets \vec{o}'_{\ACT \gets \act}), \vec{u}})$,
\commentout{
(b) $\sum_{(M,\vec{u}) \in \K} \Pr(M,\vec{u}) 
\util(w_{M, \vec{O}' \gets \vec{o}'_{\ACT \gets
    \act}), \vec{u}}) > \max_{\act' \ne \act}
\sum_{(M,\vec{u}) \in \K} \Pr(M,\vec{u}) 
\util(w_{M, \vec{O}' \gets \vec{o}'_{\ACT \gets
    \act'}), \vec{u}})$,}
and (b) $\vec{O}'$ is minimal; that is, 
for all strict subsets $\vec{O}^*$ of $\vec{O}'$, we have 
$\sum_{(M,\vec{u}) \in \K} \Pr(M,\vec{u})
\util(w_{M,\ACT \gets   \act, \vec{u}}) >
\displaystyle{\max_{\act' \in \REF(\act)} }
\displaystyle{\sum_{(M,\vec{u}) \in \K}} \Pr(M,\vec{u}) 
\util(w_{M,(\ACT \gets \act', \vec{O}^* \gets \vec{o}'_{\ACT \gets
\act}), \vec{u}})$. 
\edfn
\commentout{
either (i) $\sum_{(M,\vec{u}) \in \K} \Pr(M,\vec{u})
\util(w_{M,\ACT \gets   \act, \vec{u}}) \ge
\max_{\act' \ne \act} 
\util(w_{M,(\ACT \gets \act', \vec{O}^* \gets \vec{o}^*_{\ACT \gets \act}), \vec{u}})$,
or (ii)  $\sum_{(M,\vec{u}) \in \K} \Pr(M,\vec{u}) 
\util(w_{M, \vec{O}^* \gets \vec{o}^*_{\ACT \gets
    \act}), \vec{u}}) \le \max_{\act' \ne \act}
\sum_{(M,\vec{u}) \in \K} \Pr(M,\vec{u}) 
\util(w_{M, \vec{O}^* \gets \vec{o}^*_{\ACT \gets
    \act'}), \vec{u}})$.}

     Part (a) says that if  the variables in
$\vec{O}'$ were given the value they would get if $\act$ were performed, then
          some act $\act' \in \REF(\act)$ becomes at least as good as $\act$.
Part (b) says that $\vec{O}'$ is the minimal set of outcomes with
this property.
\fullv{
In a nutshell, $\vec{O}'$ is the minimal set of outcomes that $\ag$ is
trying to
affect by performing $\act$.  Once they have their desired values,
$\ag$ has no further motivation to perform $\act$;
some other action is at least as good.
}

What should $\REF(\act)$ be?
Since $\act \notin \REF(\act)$, if there
are only two actions in $A$, 
then $\REF(\act)$ must consist of the other act.  A natural
generalization is to take $\REF(\act) = \ACT - \{\act\}$.  The
following example shows why this will not always work.

\xam\label{xam:splitdifference}  
Suppose that Daniel is a philanthropist who is choosing a program to support.  
He wants to choose among programs that support schools and health
clinics and he cares about schools and health clinics equally.  
If he chooses program 1 he will support 5 schools and 4 clinics. 
If he chooses program 2 he will support 2 schools and 5 clinics. 
Assume he gets 1 unit of utility for each school or clinic
supported. The total utility of a program is the sum of the utility he
gets for the schools and clinics minus 1 for the overhead of both
programs.  
We can think of this overhead as the cost of implementing a program
versus not implementing any of the programs.  
By default he can also do nothing which has utility 0 since it avoids
any overhead and doesn't support any schools or clinics. 

Clearly his overall utility is maximized by choosing program 1.  
Intuitively, by doing so, he intends to affect both the schools and clinics.  
Indeed, if he could support 5 schools and 4 clinics without the
overhead of implementing a program, he would do that. 
However, if we consider all alternatives, then the minimality condition fails.
If he could support 5 schools he would switch to program 2, but if he
could support 4 clinics he would still choose program 1. This gives the
problematic result that Daniel intends to support only schools.    
The problem disappears if we take the reference set to consist of just
the default action: doing nothing.  
Then we get the desired result that Daniel intends to both support the
5 schools \emph{and} the 4 clinics.  
\exam 
\commentout{
  However, it not sufficient to just consider the default in all
  cases. Consider a slightly less scrupulous person, Jim, who can do
  nothing (the default) and his suit will be ruined and 5 people will
  die. He also has two more options. If he takes option 1 he will save
  all 5 people but his suit will still be ruined. If he takes option 2
  he will save only 1 person but his suit will stay clean. If he cares
  a little bit about lives but cares about them less than he cares
  about his suit he will choose option 2. In this case, we intuitively
  want to say he intended to keep his suit clean. However, if
  we include only the default in the reference class, we obtain the
  inference that he intended to both keep his suit clean \emph{and}
  save the 1 person.
The ``socially optimal'' action (i.e., the one that 
  optimizes the preferences of society) seems to often be included in
  the reference set.
}

\commentout{
Condition (b) by itself is also not enough to give a good definition
of intention, as the following example shows.

\xam      Suppose Louis prefers   that both Rufus and Sibella
die.  He can plant the bomb, as before.  He now has the option of
poisoning Sibella (but cannot arrange to poison Louis).  He would prefer
that Sibella die of poison rather than from the bomb, so he gets
utility 51 if just  Sibella dies, but she dies of poison and there is
no bombing; otherwise,
his utilities are the same as in the relevant variant of
Example~\ref{xam:louisintent}.  If we remove condition (a) from
Definition~\ref{dfn:intention}, then we still get that Louis intends
to kill Louis by planting the bomb, but he no longer intends to kill
both (the minimality condition for (b) fails), nor does he intend to
kill Sibella by planting the bomb (since he would prefer that she die
by poisoning). \exam
}

It might seem that by allowing $\REF(\act)$ to be a parameter of the
definition we have allowed too much flexibility, leaving room for rather
\emph{ad hoc} choices.  There are principled
reasons for restricting $\REF(\act)$ and not taking it to be all acts
other than $\act$ in general.  For one thing, the set of acts can be
large, so there are computational reasons to consider fewer acts.   
If there is a natural default
action (as in Example~\ref{xam:splitdifference}), this is often a
natural choice for $\REF(\act)$: people often just compare what they
are doing to doing nothing (or to doing what everyone expects them to
do, if that is the default).  
However, we cannot take $\REF(\act)$
to be just the 
default action if $\act$ is itself the default action (since then the
first part of Definition~\ref{dfn:intention} would not hold for any
set $\vec{O}$ of outcomes).
\shortv{We discuss the choice of reference set in more detail in the
  full paper.}
\fullv{
The choice of reference set can also be
influenced by normality considerations.  As the following example
shows, we may want $\REF(\act)$ to include what society views as the
``moral'' choice(s) in addition to the default action.

\xam\label{xam:normality} Suppose that agent $\ag$ has a choice of
saving both Tom and Jim, saving George, or doing nothing.   Saving Tom
and Jim will result in $\ag$'s nice new shoes being ruined.  Saving
either one or two 
people also incurs some minor overhead costs (it takes some time, and
perhaps $\ag$ will have to deal with some officials).  Agent $\ag$ ascribes
utility 10 to each person saved, utility $-10.5$ to ruining his shoes,
and utility $-1$ to the overhead of saving someone.  The utility of an
act is just the sum of the utilities of the relevant outcomes.  Thus,
saving Tom and Jim has utility 8.5, saving George has utility 9,
and doing nothing has utility 0.  We would like to say that by saving George,
$\ag$ intends to affect both the state of his shoes and whether George
lives or dies.  If we take $\REF(\act)$ to consist
only of the default action (doing nothing), then we get that $\ag$
intends to save George, but not that he intends to avoid ruining 
the shoes.  On the other hand, if we include the ``moral'' action
of saving Jim and Tom in $\REF(\act)$, then we get, as desired, that $\ag$
intends both to save George and to avoid ruining the shoes (whether or
not the default action is included).
\exam

The next example shows a further advantage of being able to choose the
reference set.

\xam\label{xam:defaultchoice} Suppose that $\ag$ has a choice of two
jobs.  With the first (action $\act_1$), he will earn \$1,000 and impress his
girlfriend, but not make his parents happy; with the second
($\act_2$), he will earn \$1,000 and make his parents happy, but not
impress his girlfriend; $\act_3$ correspond to not taking a job at
all, which will earn nothing, not impress his girlfriend, and not make
his parents happy.   He gets utility 5 for earning \$1,000, 3 for
impressing his girlfriend,  and 2 for making his parents happy;
there is also overhead of $-1$ in working.  Not surprisingly, he does 
$\act_1$.  If we take the default to be $\act_3$, and thus take
$\REF(\act_1) = \{\act_3\}$, then he intends both to earn \$1,000 and
impress his girlfriend.  If we take the default to be $\act_2$
(intuitively, this was the job $\ag$ was expected to take), and take
$\REF(\act_1) = \{\act_2\}$, then he intended only to impress his
girlfriend.  Intuitively, in the latter case,  we are viewing earning
\$1,000 as a given, so we do not take it to be something that the
agent intends. \exam
}

More generally, we expect it to often be the case that the reference
set consists of the actions appropriate from various viewpoints.  The
default action is typically the action of least cost, so appropriate
if one wants to minimize costs.  The socially optimal action is
arguably appropriate from the viewpoint of society.  If other
viewpoints seem reasonable, this might suggest yet other actions that
could be included in the reference set.  Of course, it may not
alwaysbe obvious what the appropriate action is from a particular
viewpoint.  As in many other problems involving causality and
reponsibility, this means that intentions are, in general, model
dependent, and there maybe disagreement about the ``right'' model.

Given the variables that the agent intends to
affect, we can determine the outcomes that the agent intends.
     
\dfn\label{dfn:intendoutcome}
Agent $\ag$ \emph{intends to bring about $\vec{O}=\vec{o}$
    in $(M,\vec{u})$ by doing action $\act$
given epistemic state $\E = (\Pr,\K,\util)$
and reference set $\REF(\act)$}
if and only if
(a) $\ag$ intended to affect $\vec{O}$ by doing action $\act$ in epistemic
state $\E$ given $\REF(\act)$, (b) there exists a setting
$(M',\vec{u}')$ such that $\Pr(M',\vec{u}') > 0$ and
\mbox{$(M',\vec{u}') \sat [\ACT \gets \act](\vec{O} = \vec{o})$,}
(c) for all values $\vec{o}^*$ of $\vec{O}$ such that there is a
setting $(M',\vec{u}')$ with $\Pr(M',\vec{u}') > 0$ and
\mbox{$(M',\vec{u}') \sat [\ACT \gets \act](\vec{O} = \vec{o}^*)$,}
we have 
$\sum_{(M,\vec{u}) \in \K} \Pr(M,\vec{u})
  \util(w_{M,\vec{O} \gets \vec{o},\vec{u}}) \ge
  \sum_{(M,\vec{u}) \in \K} \Pr(M,\vec{u}) 
  \util(w_{M,\vec{O} \gets \vec{o}^*,\vec{u}})$.
\edfn

 Part (b) of this definition says that $\ag$ considers $\vec{O} =
 \vec{o}$ a possible outcome 
of performing $\act$ (even if it doesn't happen in the actual
situation $(M,\vec{u})$).  Part (c) says that, among all possible
values of $\vec{O}$ that $\ag$ considers possible, $\vec{o}$ gives the
highest expected utility.

This definition seems to capture significant aspects of natural
language usage of the word ``intends'', at least if $\act$ is
deterministic or close to deterministic.  But 
if $\act$ is probabilistic, then we often use the word ``hopes''
rather than intends.  It seems strange to say that $\ag$ intends to
win \$5,000,000 when he buys a lottery ticket (if
\$5,000,000 is the highest payoff); it seems more reasonable to say
that he hopes to win \$5,000,000.    Similarly, if a doctor performs
an operation on a patient who has cancer that he believes has only a
30\% chance of complete remission, it seems strange to say that he
``intends'' to cure the patient, although he certainly hopes to cure
the patient by performing the operation.  In addition, once we think
in terms of  
``hopes'' rather than ``intends'', it may make sense to consider not
just the best outcome, but all reasonably good outcomes.  For example,
the agent who buys a lottery ticket might be happy to win any prize
that gives over \$10,000, and the doctor might also be happy if the
patient gets a remission for 10 years.

\commentout{
    >     (b) and (c) are parts (b) and (e) of the current definition.  I've
    >     replaced the current (c) and (d) by (a).  Part (e) says that \vec{o}
    >     is the best possible value of \vec{O}.  That's fine if \vec{O} is a
    >     single binary variable, but it seems too strong.  If you're a doctor
    >     performing an operation with many possible outcomes, which ones do you
    >     intend?   If a pretty good outcome happens that's not the best, do we
    >     really want to say that you don't intend that?  I would say that we
    >     intend the union of all outcomes that are pretty good (say, better
    >     than the expected outcome we can get if we perform another action).
    >     So I would change (e) too.
    >

\dfn\label{dfn:intention} An agent $\ag$ 
\emph{intended to bring about $\vec{O}=\vec{o}$
  in $(M,\vec{u})$ by doing $\act$
  given epistemic state $\E = (\Pr,\K,\util)$} if and only if
there
exists an extension $\vec{O}' = \vec{o}'$ of $\vec{O} = \vec{o}$
that does not include a conjunct of the form $\ACT = \act''$,
an action $\act'$, and $(M',\vec{u}') \in \K$
such that
\commentout{
(a) $(M,\vec{u}) \sat \ACT = \act \land \vec{O}' = \vec{o}'$, (b)
$$
\begin{array}{l}
\sum_{(M,\vec{u}) \in \K} \Pr(M,\vec{u}) 
(\util(w_{M,A \gets \act', \vec{O}' \gets \vec{o}',\vec{u}}) -
\\ \qquad \qquad \qquad \util(w_{M,A \gets \act, \vec{O}' \gets \vec{o}',\vec{u}}))
\ge 0,\end{array}$$
and (c) for all strict subsets $\vec{O}^*$ of $\vec{O}'$, all
actions $\act'$, if $\vec{o}^*$ is the restriction of $\vec{o}^*$, we have
variables in $\vec{O}^*$, then
     $$
     \begin{array}{l}
     \sum_{(M,\vec{u}) \in \K} \Pr(M,\vec{u}) 
(\util(w_{M,A \gets \act, \vec{O}^* \gets \vec{o}^*,\vec{u}}) -
     \\ \qquad \qquad \qquad \util(w_{M,A \gets \act', \vec{O}^* \gets \vec{o}^*,\vec{u}}))
     >  0.\end{array}$$
 \edfn
}
(a) $(M,\vec{u}) \sat \ACT = \act$, (b) $\Pr(M',\vec{u}') > 0$ and
$(M',\vec{u}') \sat [\ACT \gets \act](\vec{O} = \vec{o})$, 
(c) $
\sum_{(M,\vec{u}) \in \K} \Pr(M,\vec{u}) 
(\util(w_{M,A \gets \act', \vec{O}' \gets \vec{o}',\vec{u}}) -
\util(w_{M,A \gets \act, \vec{O}' \gets \vec{o}',\vec{u}}))
\ge 0$,
(d) for all strict subsets $\vec{O}^*$ of $\vec{O}'$, all
actions $\act'$, if $\vec{o}^*$ is the restriction of $\vec{o}'$ to
the variables in $\vec{O}^*$, then
$     \sum_{(M,\vec{u}) \in \K} \Pr(M,\vec{u}) 
(\util(w_{M,A \gets \act, \vec{O}^* \gets \vec{o}^*,\vec{u}}) -
\util(w_{M,A \gets \act', \vec{O}^* \gets \vec{o}^*,\vec{u}}))
     >  0$,
     and (e) for all $(M'',\vec{u}'')$ such that $\Pr(M'',\vec{u}'')
     > 0$, we have $u(w_{M, \vec{O}' \gets \vec{o}',\vec{u}''}) \ge u(w_{M,\vec{u}''})$.
       \edfn

       Part (a) of this definition just says that for that $\ag$ to intend an
outcome by performing $\act$, he must actually perform $\act$.  Part
(b) says that $\ag$ considers $\vec{O} = \vec{o}$ a possible outcome
of performing $\act$ (even if it doesn't happen in the actual
situation $(M,\vec{u})$).  Part (c) says that there is a superset 
$\vec{O}'$ of $\vec{O}$ that consists of the variables that $\ag$
wants to affect by doing $\act$; if these variables already have their
desired value $\vec{o}'$, then $\act$ would not bother doing $\act$.
Part (d) says that $\vec{O}'$ is minimal; $\ag$ would continue to do
$\act$ if only a strict subset of the variables in $\vec{O}'$ was
determined.  Finally, part (e) says that the values $\vec{o}'$ are the
best values for the variables in $\vec{O}'$; these are the values
$\ag$ would like the variables in $\vec{O}'$ to take on (although they
might not take on these values in the actual world, as is the case in
Example~\ref{xam:doctor}).  
}

\commentout{
Note that if $\ag$ intended to bring about $\vec{O} =
\vec{o}$ by performing $\act$, then he intended to perform $\act$, in
the sense of Definition~\ref{dfn:accident}.
This follows
by  simply taking $\vec{O}^* = \emptyset$ in condition (d) above.
}


%

\fullv{
\xam\label{xam:trolley2} In the basic trolley scenario,
   under minimal assumptions (namely, that the agent's utility
   function is such that fewer deaths are better), an agent who pulls
   the lever does not intend to kill the person on the side track;
   he would have pulled the lever 
   even if the person did not die (since the train would still have
   gone on the side track, and the 5 deaths would have been avoided).
   The situation is a bit more subtle in the case of the loop
      problem.  What the agent intends depends in part on how we model
   the problem.  One reasonable way is to have a binary variable $\TrH$ for
   ``trolley hits the person on the side track'', a binary variable $D$
   for ``person on side track dies'', and a binary variable $\TS$ for
   ``train stops before killing the 5 people on the main track''.  We
      have the obvious equations: $D = \TrH$ and $\TS = \TrH$;
   as a result of the train hitting the person on the side track, he
   dies and the train stops before hitting the 5 people on the main
   track.
 It     follows from Definition~\ref{dfn:intention} that the
   agent $\ag$ who pulled the 
   lever intended to hit the person
on the side track, but did     not intend to kill him.


   Note that if we do not separate the train hitting
   the person on the track from the death of that person, but rather
   just have the variables $D$ and $\TS$, with the equation $\TS = D$,
   then $\ag$ did intend the person's death.
   Given that the death causes the train to stop, not pulling the
   lever is at least as good as pulling the lever (either way, the
   train does not hit the five people), and slightly better if there
   is a small cost to pulling the lever.
   Intuitively, the latter choice of model is appropriate if the
   person cannot even imagine hitting the person without killing him.
   This shows that 
      intention (just like causality and blameworthiness) is
   very much model dependent, and, among other things, depends on the
   choice of  variables.

The issues that arise in choosing the variables in a causal model
arise more generally in causal
   chains where both means and ends are desired. 
   \xam\label{xam:study} Consider a student $\ag$ that can decide
      whether or not to study ($\ACT = s$ or $\ACT = ns$) for an exam. If
      $\ag$ studies she will 
   receive good grades ($G$) and get a job ($J$) otherwise she will
   receive poor grades and not get a job. Assume the cost of studying
   is $-1$ to $\ag$ but the value of good grades is 10 and the value of a
   job is also 10. Thus, for $\ag$, either the grades or job would have
   been sufficiently motivating to study. Intuitively, we would like to
   say that $\ag$ intends to both get good grades and get a
  job. However, according to our definition of intention, in this
  model, the agent does not
  intend to get a job.  Setting $J$ to 1 is by itself not enough for
  the agent not to study, since he also wants to get good grades.
  While setting both $J$ and $G$ to 1 is enough for the agent not to
  study, this is not a minimal intervention; setting $G$ to 1 is
  enough to get the agent not to study (since setting $G$ to 1 causes
  $J$ to be 1).  Like the loop track case, if we augment the model to
  include a variable $A$ representing the sense of accomplishment that
  $\ag$ feels as a result of getting good grades, with the obvious
  equation $A = G$, then in the
  resulting model, $\ag$ intends to both get good grades and get a job
  (since setting $A$ and $J$ to 1 suffices to get the agent not to
  study, and this is a minimal set).  The variable $A$ plays the same
  role here as the variable $D$ in the model of trolley problem with
  the loop; it enables us to separate the means---getting good
  grades---from the end---the sense of accomplishment.
  Once we use different variables for the means and ends in this way,
  we can examine more carefully what the agent truly intended.%
  \footnote{We thank Sander Beckers for suggesting this example and
  pointing out in the original model, the agent does not intend to get
  a job.}
\exam
   
   \commentout{
   In police stories we often hear murderers
   say things like ``I didn't intend to kill him; I just wanted to
   hurt him enough to give him a warning \ldots''  The murderer here
   is appealing to the same intuitions that suggest adding the
   variable $\TrH$ to the model above.}

      If we change the trolley problem so that the person on the side
track is planning on blowing up 10 other people, 
then according to our definition, the
agent who pulls the lever intends to both kill the  person on
the side track \emph{and} to save the five people on the main track.  Our
definition delivers the desired result here.
\exam
}


\commentout{
   As the following example shows, it is possible that $\ag$ is happy
      that outcome $\phi$ occurred, could have done something to prevent it,
   but yet did not intend for it to happen.

   \xam
   Suppose that Bob does not like Charlie and knows that Charlie does
   not like really spicy food.   Bob notices Charlie, who is sitting
   two tables away at a restaurant,  and is about to eat a dish that
   Bob knows to be extremely spicy.  Bob says nothing, happy that 
   Charlie will feel uncomfortable.  Does Bob intend Charlie's
   discomfort?  According to Definition~\ref{dfn:intention}, he does
   not.
   \mtodo{I'm not sure I understand this... my intuition and
     understanding of the definition is that he does intend this
     (especially without the disutility on speaking across the
     restaurant).}
   Assuming that there is a small disutility in shouting across
   the restaurant, Bob would have continued to say nothing even if the
   spicy food had not affected Charlie at all. \mtodo{right -- so we
     should say he intended to not violate the speaking norm AND for
     Charlie to get a stomach ache. Isn't this just the classic
     over-determination. An agent has to have an intended outcome of
     their action. }
   Note that this is
   different from Bob deliberately adding a lot of spice to Charlie's
   meal. If we assume that there is a small disutility to Bob for
   doing so, then in a context where he adds the spice, Bob does
   intend Charlie's disutility; he presumably would not add it if
   Charlie would not feel discomfort.
   \mtodo{yes this seems true since its the standard side-track model (one pleasant and one unpleasant outcome)}
   \exam
}

\fullv{
The following example, which is due to Chisholm \citeyear{Chis66}
and discussed at length by Searle \citeyear{Searle83}, has been
difficult for other notions of intention to deal with, but is not a
problem for the definition above.

\xam\label{xam:Chisholm}  Louis wants to kill his uncle and has a
plan for doing so.  On the way
to his uncle's house in order to carry out his plan, he gets so
agitated due to thinking about the plan that he loses control of his 
car, running over a pedestrian, who turns out to be his uncle.
Although Louis wants to kill his uncle, we would not want to say
that Louis intended to kill his uncle by running over the pedestrian,
nor that he intended to run over the pedestrian at all.  Given
reasonable assumptions about Louis's beliefs (specifically, that the
pedestrian was extremely unlikely to be his uncle), he clearly would
have preferred not to run over the pedestrian than to run him over, so
the action of running over the pedestrian was not intended according to
Definition~\ref{dfn:accident}.  Thus, he did not intend his uncle to
die when he ran over over the pedestrian.
\exam
}

\commentout{
\section{Putting it all together}\label{sec:moralresp}
People's ascriptions of moral responsibility seem to involve three
components that we have called here causality, degree of 
blameworthiness, and intention.%
\footnote{In his influential work, Weiner \citeyear{Weiner95}
  distinguishes causality, responsibility, and blame.  Responsibility
  corresponds roughly to what we have called blameworthiness, while
  blame roughly corresponds to blameworthiness together with
  intention.}
If $\ACT=\act$ is not a cause of
$\phi$, then the agent $\ag$ who performed $\act$ is not morally responsible
for $\phi$ at all.  If $\ACT=\act$ is a  cause of $\phi$, then the
degree to which $\ag$ is responsible for $\phi$ depends on some
combination of intention and degree of blameworthiness.
\commentout{
Retrospectively, that is, after the fact, we can ask the extent to
which $\ag$'s action $\act$ was morally responsible for outcome $\phi$.
For $\act$ to be morally responsible for $\phi$, it is necessary for
$\ACT = \act$ to be a cause of $\phi$ according  to
Definition~\ref{actcaus} (or some other reasonable substitute).  This,
in turn, implies that $\act$ was actually performed and $\phi$ actually
happened.  If $\ACT = \act$ is part of a cause of $\phi$, we can consider both
the degree of moral responsibility of $\ACT = \act$ for $\phi$ given
epistemic state $\E$ (according to Definition~\ref{dfn:moralresp}) and
(if $\phi$ is of the form $O=o$) whether $\ag$ intended to bring about
$\phi$.
}
\commentout{
  People seem to consider both degree of blameworthiness and intention
when determining degree of moral responsibility.
Results of Mikhail
\citeyear{Mikhail07} show that people judge pulling the lever less
morally permissible in the loop condition than in the basic side track
condition of the trolley problem; results of Kleiman-Weiner et
al.~\citeyear{KWG15} suggest that this is due to the fact that 
the agent who pulls the lever intends to hit the man in the loop
condition, but not in the basic version of the trolley problem.
The extent to which people consider intention is strongly affected by
how much physical contact there is (e.g., there seems to be a difference between
pushing someone onto the track to stop the train and just pulling a
lever), the degree of anonymity of the people involved, and the mood
of the people making the judgment (e.g., whether they had just watched
a comedy or a boring documentary).  (See \cite[Section 3]{Edmonds14}
for more discussion of these issues and further references.)  }
Because it is not clear exactly how intention and degree of blame
should be combined, we have left them here as separate components of
moral responsibility.
}


   %

\commentout{
In addition to moral responsibility, people consider ``moral permissibility''.
Blameworthiness and moral responsibility are defined
for an action
relative to an
outcome.  By way of contrast, we talk about actions being morally
permissible without talking about specific outcomes.  As a first cut,
we can view
action $\act$ as morally permissible if there exists
a ``reasonable'' epistemic state $\E$ such that $\act$ is the action that
gives highest expected utility with respect to $\E$.  This definition
leaves a lot of flexibility as to what counts as morally permissible,
since people can certainly disagree on what counts as a reasonable
epistemic state.  Our society would not consider Louis's utility function in
Example~\ref{xam:louisintent}, which gives high utility to Rufus dying,
to be reasonable; but people might find an attempt to assassinate a horrible
dictator by planting a bomb at the table where he is dining morally
permissible, and might continue to judge it so even if some innocent
people died as well.  We do not judge young children harshly become we
are willing to consider a wider range of probabilities and utilities
as acceptable for them.
Moreover, in the trolley problem, most people would consider it acceptable
that a person should ascribe higher utility to saving, say, his
wife than to saving a random person.  

This is only a first cut at a definition of moral permissibility
because, as results of Mikhail \citeyear{Mikhail07} show, 
intention also seems to factor into moral permissibility judgments.
Some people seem to ``downgrade'' the moral permissibility of an
action if it resulted in a bad outcome that the agent who performed
the action intended, even if the overall utility of the action is high.
Specifically, Mikhail
shows that people judge pulling the lever less
morally permissible in the loop condition than in the basic side track
condition of the trolley problem. Results of Kleiman-Weiner et
al.~\citeyear{KWG15} suggest that this is due to the fact that 
the agent who pulls the lever intends to hit the man in the loop
condition, but not in the basic version of the trolley problem.
However, 
the extent to which people consider intention seems to be strongly affected by
how much physical contact there is (e.g., there seems to be a difference between
pushing someone onto the track to stop the train and just pulling a
lever), the degree of anonymity of the people involved, and the mood
of the people making the judgment (e.g., whether they had just watched
a comedy or a boring documentary).  (See \cite[Section 3]{Edmonds14}
for more discussion of these issues and further references.)  
}

\fullv{
Experiments performed by Kleiman-Weiner et al.~\citeyear{KWG15} lend support to
  the fact that people are using utility considerations in judging
degree of moral permissibility.
For example, the more people there are on the main
track, the greater the number of people who judge it morally
permissible to pull the lever.  Presumably, pulling the lever has
greater utility if it saves more people.  In addition, in a situation
where there is only one person on the main track, but it is $\ag$'s
brother, it is considered more morally permissible to pull the lever
than if the one person on the main track is an anonymous individual.
Presumably, $\ag$ gets higher utility by saving his brother than by
saving an anonymous person; people considering moral
responsibility take that into account.}
\fullv{
Kleiman-Weiner et al. \citeyear{KWGprep} provide a
theory of moral permissibility which generalizes the doctrine of
double effect\footnote{The \emph{doctrine of double effect} is a well studied
moral rules that says that an action is permissible if the
  agent performing that action intends the good effects and does not
  intend the bad effects as either an end or as a means to an end. The
  good effects must also outweigh the negative unintended
  side-effects. } and   
integrates both utility maximization and intention using a noisy-or model. 
}


\commentout{
Both blameworthiness and intention are defined with respect to an
epistemic state $\E$.   However, we may not want to use the same
epistemic state for both.  As discussed above, for
blameworthiness, we typically want to use the epistemic state of a
``reasonable person''.
On the other hand, in
the definition of intention, we must use the agent's actual
epistemic state.
}


\fullv{
The three components that make up moral responsibility involve a mix
of retrospective and prospective judgments.  Causality is purely
retrospective; it is based on what happened.  Blameworthiness
as we have defined it
is purely
prospective; it is based on beliefs that hold before the action was
performed.  The notion of ``$\act$ was intended'' given in
Definition~\ref{dfn:accident} is retrospective; we don't say that
$\act$ was (un)intended unless $\act$ was actually performed.  On the
other hand, the notion of ``intending to bring about $\vec{O} =
\vec{o}$ by doing $\act$'' is prospective.  
When an autonomous agent is applying these
definitions, we would expect the focus to be on the prospective parts,
especially blameworthiness.
Interestingly, Cushman \citeyear{Cushman08} shows that people
distinguish between how ``wrong'' an action is (which depends largely
on the agent's mental state, and what the agent believed) and whether an agent
should be punishment for the action (which depends on 
causality---what actually happened---as well as the agent's epistemic
state).  Both notions depend on 
intention.  The point is that whether an act is right or wrong is
essentially prospective, while whether someone should be punished has
both prospective and retrospective features.  Although we use words
like ``blame'' both in the context of right/wrong and in the context
of ``deserving of punishment'', Cushman's work shows that people are
quite sensitive to the prospective/retrospective distinction.  Thus,
it seems useful to have formal notions that are also sensitive to this
distinction.
}

\commentout{
When we consider these definitions prospectively, for example, when an
autonomous agent is trying to decide whether to perform an action
$\act$, we cannot determine causality, but we can still use the
epistemic state to determine
(a) the probability that $\phi$ will happen if action $\act$ is
performed, (b) whether the agent intends $\phi$ if $\act$ is
performed, and (c) what the agent's degree of moral responsibility
would be for $\phi$ if $\act$ were performed and ended up causing the
$\phi$.%
\footnote{Here it is particularly important that $\Pr$ represents the
probability before the action is performed, so it is a prospective
evaluation in any case.}  This is the information that the autonomous
agent would need to help it evaluate the decision.
}


\fullv{
We conclude this section by considering perhaps the best-studied example
  in the moral responsibility literature, due to Frankfurt
\citeyear{Frankfurt69} (see,  e.g., \cite{WM03} for more discussion of
this example).

\xam\label{xam:frankfurt}
Suppose that Black wants Jones
to kill Smith, and is prepared to go to considerable lengths to ensure
this.  So Black waits to see if Jones poisons Smith's drink.  If Jones
does not do this, Smith would give Jones a loaded gun and persuade him
to kill Smith anyway.  
We are supposed to assume here that Smith can tell if Jones put a
poison in Smith's drink and can persuade Jones to shoot Smith.
However, Jones is uncertain about Black's intent and his persuasive powers.
(Indeed, in many variants of the story, Jones does not even know that
Black is present.)
In any case, Jones does in fact 
poison Smith, and consequently Smith dies.

The problem for most theories of moral responsibility here is that,
although Jones freely chose to poison Smith, there
is a sense in which he could not have prevented himself from being a cause 
of Smith's death, because had he not poisoned Smith, Black would have
persuaded Jones to shoot Smith.

Despite Black, Jones' action of poisoning is a cause of
Smith's death 
according to Definition~\ref{actcaus}. If we consider the obvious
causal model $M$ with with an exogenous variable $\JP$ (Jones poisons Smith)
and endogenous variables $\BP$ (Black persuades Jones to
shoot Smith), $\JS$ (Jones shoots Smith), and
$\SD$ (Smith dies), with the obvious equations ($\SD = \JS \lor \JP$, 
$\JS = \BP$, $\BP = \neg \JP$, and $u$ is a context where $\JP = 1$, 
then $$(M,u) \sat \JP = 1
\land \BP = 0 \land \JS = 0 \land \SD = 1.$$  Since
$(M,u) \sat [\JP \gets 0, \BP \gets 0](\SD = 0)$, it follows that
$\JP = 1$ is a cause of Smith's death.  Moreover, if Jones in fact poisons
Smith, it seems reasonable to assume that his utility function is such
that he intended the poisoning and its outcome.  Jones has a
positive
degree of blameworthiness for Smith's death if we assume
that $\Pr$ assigns positive probability to a causal model where Jones
poisons Smith and Black would not be able to persuade Jones to shoot
Smith if Jones didn't poison him, either because he didn't bother
trying or his persuasive powers were insufficient.  
Interestingly, if $\Pr$ assigns probability 1 to Black wanting to and
being able to persuade Jones, then Jones will have degree of 
blameworthiness 0 for Smith's death, although he intends to kill him.
The lower the probability of Black wanting to and being able to
persuade Jones, the higher Jones' blameworthiness, and hence the higher
the Jones' moral responsibility.

This seems to us reasonable.  Consider the following more realistic
Frankfurt-style problem.  Jones votes for Smith in an election.  The
army wants Smith to win, and if he does not win, they will step in and
declare him the victor.  If the probability of the army being able to
do this is 1, then it seems reasonable to say that Jones has 
degree of blameworthiness 0 for the outcome.  On the other hand, it seems
unreasonable to say that the army is certain to be able to
install Smith even if he does not win.  We cannot be certain of this
outcome, although it may seem reasonable to give it high probability.
The higher the probability, the lower Smith's degree of
blameworthiness. \exam
}

\section{Complexity considerations}\label{sec:complexity}
Since $w_{M,\vec{X}=\vec{x},\vec{u}}$ can be computed in time
polynomial in the size of $M$, it easily follows that, given an
epistemic state $\E = (\Pr,\K,\util)$, 
$\delta_{\act,\act',\phi}$ can be computed in time polynomial in
$|\K|$.  Thus, the degree of blameworthiness of an action $\act$ for
outcome $\phi$ can be computed in time polynomial in $|\K|$ and the
cardinality of the range of $\ACT$.  Similarly, whether $\act$
is (un)intended in $(M,\vec{u})$ given $\E$ can be computed in time
polynomial in $|\K|$ and the cardinality of the range of $\ACT$.

The complexity of determining whether $\ACT=\act$ is part of a cause of
$\phi$ in $(M,\vec{u})$ is $\Sigma_2^p$-complete,  that is,
it is at the second level of the polynomial hierarchy \cite{Sipser96}.
This complexity is due to the ``there exists--for all'' structure of
the problem (there exist sets $\vec{X}$ and $\vec{W}$ of variables
such that for all strict subsets of $\vec{X}$ \ldots).  The problem of
determining if $\ag$ intended to bring about $\vec{O}=\vec{o}$ has a similar
``there exists--for all'' structure; we conjecture that it is also
$\Sigma_2^P$-complete.  While this makes the general problem quite
intractable, in practice, things may not be so bad.  Recall that
$\ag$ intends to bring about $\vec{O} = \vec{o}$ if there exists a
superset $\vec{O}'$ of $\vec{O}$
(intuitively, all the outcomes that $\ag$ intends to affect)
with the appropriate properties.
In practice, there are not that many outcomes that determine an
agent's utility. If we assume  that $|\vec{O}'| \le k$ for some fixed $k$, 
then the problem becomes polynomial in the number
of variables in the model and the number of actions; moreover, the 
polynomial has degree $k$.  In practical applications, it seems reasonable 
to assume that there exists a (relatively small) $k$, making the
problem tractable.

\section{Related work}\label{sec:related}
Amazon lists over 50 books in Philosophy, Law, and Psychology with the
term ``Moral Responsibility'' in the title, all of which address the types
of issues discussed in this paper.  There are dozens of other books on
intention.
Moreover, there are AI systems that try to build in
notions of moral responsibility (see, e.g., \cite{DTFK08,MG12,SMB15}).
Nevertheless, there has been
surprisingly little work on providing a formal definition of moral
responsibility of the type discussed here.
\fullv{
  Although other authors have used models that involve
probability and utility (see below), we are not aware of any formal
definition of degree of blameworthiness.}
We now briefly discuss some of the work most
relevant to this project, without attempting to do a comprehensive
survey of the relevant literature.
\shortv{We go into more detail in the full paper.}

As mentioned in the introduction, Chockler and
Halpern~\citeyear{ChocklerH03} define a notion of responsibility that 
tries to capture the diffusion of responsibility when multiple
agents contribute to an outcome but no agent is a \emph{but-for} cause of
that outcome, that is, no agent can change the outcome by just
switching to a different action.
For example, the degree of responsibility of a
voter for the outcome $1/(1 + k)$, where $k$ is the number of changes
needed to make the vote critical (i.e., a but-for cause).
\fullv{For example, in a 6--5 vote, each of the 6 voters
who voted for the outcome has degree of responsibility 1, since they
are all critical; if anyone changes her vote, the outcome will be
different.  In an 11--0 vote, each voter has a degree of
responsibility of $1/6$, because 5 other votes need to flip to make
the vote 6--5, at which point that voter is critical. } Chockler and
Halpern also use epistemic states (although without the utility
component):  they define a notion of \emph{degree
  of blame} given an epistemic state $\E$, which is the expected
degree of responsibility with respect $\E$. These notions of blame and
responsibility do not take utility into account, nor do they consider
potential alternative actions or intention.

Cohen and Levesque \citeyear{CL90} initiated a great deal of work in
AI on reasoning about an agent goals and intentions.
They define a modal logic that includes operators for goals and
beliefs, and define formulas $\INTEND_1(\ag, \act)$---agent $\ag$
intends  action $a$---and $\INTEND_2(\ag,p)$---agent $\ag$ intends goal
$p$.   $\INTEND_1(\ag,\act)$ is the analogue of 
Definition~\ref{dfn:accident}, while
$\INTEND_2(\ag,p)$ is the analogue of Definition~\ref{dfn:intendoutcome}; a
goal for Cohen and Levesque is essentially an outcome.
Roughly speaking,
agent $\ag$ intends to bring about $\phi$ if $\ag$ has a plan that he
believes will bring about $\phi$ (belief is captured using a modal
operator, but we can think of it as corresponding to ``with high
probability''), is justified in believing so, and did not intend to
bring out $\neg \phi$ prior to executing the plan.
\fullv{
(Cohen and Levesque need 
the latter condition to deal with examples like
Example~\ref{xam:Chisholm}.)}
Their framework does not allow us to model an agent's utility, nor can
they express counterfactuals.
\fullv{Part of the reason that Cohen and
Levesque have to work so hard is that they consider plans over time
(and their definition must ensure that $\ag$ remains committed to the plan);
another difficulty comes from the fact that they do not have an easy way to
express counterfactuals in their model.
}

\fullv{Van de Poel et al.~\citeyear{PRZ15} focus on what they  call
  \emph{the problem 
    of many hands} (a term originally due to Thompson \citeyear{Thompson80}): that is, the problem of allocating responsibility
  to indivdual agents who are members of a group that is clearly
    responsible for an outcome.  This is essentially the problem noted
in the discussion of overfishing after Example~\ref{xam:tragedy}.
They consider both prospective and retrospective notions of
responsibility.   They formalize some of their ideas using a variant
of the logic CEDL (\emph{coalition epistemic dynamic logic})
\cite{LR15}.
Unfortunately, CEDL cannot directly capture counterfactuals, nor can
it express quantitative notions like probability.  Thus, while it can
express but-for causality, it cannot capture most of the more subtle
examples of causality, such as the Billy-Suzy rock-throwing example discussed
in Section~\ref{sec:causality}, nor can it capture more quantitative
tradeoffs between choices that arise when defining degree of
blameworthiness.} 

\fullv{Gaudini, Lorini, and Mayor \citeyear{GLM13} 
discuss moral responsibility for a group, but do not
discuss it for a single agent. It is not obvious how their approach
would be applied to a single-agent setting. They have no causal
model, instead considering a game-theoretic setting where agents are
characterized by their guilt-aversion level.  It does seem
that, with many agents, game-theoretic concerns should be relevant,
although we believe that a causal model will be needed as well.   It
would be of interest to consider an approach that combines both
causality and game theory to analyze moral responsibility for a group.

Lorini, Longin, and Mayor \citeyear{LLM14}  use STIT (``seeing to it
that'') logic, which was
previously used by Lorini and Schwarzentruber \citeyear{LoriniS10} to capture
causality.  The logic includes operators that can express notions like
``group $J$ can see to it that $\phi$ will occur, not matter what the
agents outside $J$ do''.  The STIT logic lacks counterfactuals, so it will have
difficulty dealing with some of standard examples in the causality
literature (see \cite{Hal48}).   Lorini, Longin, and Mayor also focus
on group notions, such as collective responsibility.

Barreby, Bourgne, and Ganascia \citeyear{BBG15} also provide a logic
for reasoning about moral responsibility based on the event calculus
\cite{KS86},
and show how it can be implemented using answer set programming
\cite{Gelfond08}. 
They show how the trolley problem can be captured using their
approach.   Like the STIT approach, their version of the event
calculus cannot express counterfactuals, so we again do not believe
that their approach will 
be able to capture adequately the causal issues critical to reasoning
about moral responsibility. 
}

Kleiman-Weiner et al. 
\fullv{\citeyear{KWG15,KWGprep}} \shortv{\citeyear{KWG15}} give a definition of   
intention in the spirit of that given here.  Specifically, it involves
counterfactual reasoning and takes expected utility into account.  
It gets the same results for intention
in the standard examples
as the definition given
here, for essentially the same reasons. However, rather than using
causal models, they use influence diagrams.
 The agent's intention when performing $\act$ is then a minimal set of
  nodes whose fixation in the influence diagram would result in some 
    action $\act'$ having expected utility at least as high as that of
    $\act$.
\shortv{ They use their definition of intention along with a model for the
    inference of the agent's epistemic state and utility function to
        model human judgments in many moral dilemmas.} 
    \fullv{Kleiman-Weiner et al.~also build on this model to give
    a theory of moral permissibility which generalizes the doctrine of double effect and 
    integrates both utility maximization and intention \cite{KWGprep}.
    Their model is tested against human judgments across many moral dilemmas.}

\fullv{
Vallentyne \citeyear{Vallentyne08} sketches a theory of moral responsibility
that involves probability.  Specifically, he considers 
the probability of each outcome, and how it changes
as the result of an agent's choice, without using utility and taking
expectation. Thus, he works with tuples of probabilities (one for each
outcome of interest).  Rather than using counterfactuals,  he takes
$A$ to be a cause of $B$ if performing $A$ raises the probability of
$B$.%
\footnote{This approach to causality is known to not deal well with
many examples;   see \cite{Hal48}.}
Thus, an agent is responsible for an outcome only if his action raises 
the probability of that outcome.  His model also takes into account
the probability of an agent's disposition to act.  This seems hard to
determine.  Moreover, while Vallentyne uses disposition as an input to
determining moral responsibility; for autonomous agents, we would want
the agent's disposition to depend in part on moral responsibility.

}

Perhaps closest to this paper is the work of Braham and van Hees
\citeyear{BH12}.   They say that an agent $\ag$ is morally
responsible for an outcome $\phi$ if (a) his action $\act$ was a cause of 
$\phi$, (b) $\ag$ intended to perform $\act$, and (c) $\ag$ had no 
eligible action $\act'$  with a higher \emph{avoidance
  potential}.
\fullv{They define cause using Wright's \citeyear{wright:88}
  notion of a \emph{NESS test} (Necessary Element of a Sufficient Set),%
\footnote{See \cite{Hal39} for a discussion of problems with using the
  NESS test to define causality.} and}
\shortv{They} do not give a formal definition of intentionality, instead
assuming that in situations of interest to them, it is always
satisfied.
Roughly speaking, the avoidance
potential of $\act$ with respect to $\phi$ is the probability that 
$\act$ does not result in $\phi$.
Thus, the notion of the avoidance potential of an action $\act$ for
$\phi$ being greater than that of $\act'$ is somewhat related to
having
$\delta_{\act,\act',\phi} > 0$, although the technical details are
quite different.
\fullv{
Braham and van Rees consider a
multi-agent setting, where all the uncertainty is due to
uncertainty about what the other agents will do; they further assume
  that the outcome is completely determined given a strategy for each
  agent (so, in particular, in the single-agent case, their setting is
  completely deterministic; they do not allow uncertainty about the
  outcome).%
  \footnote{Since Braham and van Rees do not make use of any of the
    machinery of game theory---in particular, for them, the
    probabilities of other agents' strategies do not necessarily arise
    from equilibrium considerations---there is no difficulty in
    identifying a strategy profile (a description of the strategy used
    by each of the agents) with a context, so having a probability on
    contexts as we have done here is more general than having a
    probability on other agents' strategies.}
  Nevertheless, it is clear that their
  notion of avoidance potential is trying to compare outcomes of
  $\act$ to those of other acts, in the spirit of
  Definition~\ref{dfn:moralresp}.  
}

\shortv{In the full paper, other recent work
  \cite{BBG15,GLM13,LLM14,PRZ15,Vallentyne08} is also discussed.}

\section{Conclusion}\label{sec:conclusion}

People's ascriptions of moral responsibility seem to involve three
components that we have called here causality, degree of 
blameworthiness, and intention.
We have given formal definitions of the latter two.
Because it is not clear exactly how intention and degree of blame
should be combined, we have left them here as separate components of
moral responsibility.%
\footnote{In his influential work, Weiner \citeyear{Weiner95}
  distinguishes causality, responsibility, and blame.  Responsibility
  corresponds roughly to what we have called blameworthiness, while
  blame roughly corresponds to blameworthiness together with
  intention.}
Considerations of moral responsibility have become more
pressing as we develop driverless cars, robots that will help in
nursing homes, and software assistants.  The framework
presented here should help in that regard.

Our definitions of blameworthiness and intention were given relative
to an epistemic state that 
included a probability measure and a utility function.  This means
that actions could be compared in terms of expected utility; this
played a key role in the definitions.  But there are some obvious
concerns: first, agents do not ``have'' complete probability
measures and utility functions.  Constructing them requires nontrivial
computational effort.  Things get even worse if we try to consider
what the probability and utility of a ``reasonable'' person should be;
there
will clearly be far from complete agreement about what these should be.
And even if we could agree on a probability and utility, it is not
clear that maximizing expected utility is the ``right'' decision rule.
One direction for further research is to consider how the
definitions given here play out if we use, for example, a set of
probability measures rather than a single one, and/or use decision rules
other than expected utility maximization (e.g., maximin).
Another issue that deserves further investigation is
responsibility as a member of the group vs. responsibility as an
individual (see the brief discussion after
Example~\ref{xam:tragedy}).

One final comment: the way we have used ``blameworthy''
in this paper is perhaps closer to the way others might use the word
``responsible''.  That is, some people might say that we should not
blame the person who killed one rather than 5 in the trolley problem,
although that person is definitely responsible for the one death.  
There is a general problem in this area that English tends to use a
small set of words (``blame'', ``responsibility'', ``culpability'')
for a complex of closely related notions.  People are typically not
careful to distinguish which notion they mean.  One advantage of
causal models is that they allow us to tease apart various notions,
such as what we have called ``blameworthiness'' here and the notions 
of  ``responsibility'' and ``blame'' as defined by Chockler and
Halpern \citeyear{ChocklerH03}.  There may be other related notions
worth considering.  We hope that the particular words we have chosen
to denote these notions does not confuse what we consider the
important underlying issues.

\paragraph{Acknowledgments:}
Halpern was supported in part by NSF 
grants IIS-1703846 and IIS-1718108, AFOSR grant FA9550-12-1-0040, 
ARO grant W911NF-17-1-0592, and a grant from the Open Philanthropy
project.  Kleiman-Weiner was supported in part 
by a Hertz Foundation Fellowship.
We thank Sander Beckers, Tobias
Gerstenberg and Jonathan Phillips for interesting discussions and many useful comments on
the topics of this paper.  

\bibliographystyle{aaai}
\bibliography{z,joe}

   \end{document}